\documentclass{bmvc2k}

\usepackage{times}
\usepackage{epsfig}
\usepackage{graphicx}
\usepackage[xvgnames]{xcolor}
\definecolor{ocre}{RGB}{0, 0, 64}
\usepackage[font={color=ocre,small},labelfont=bf]{caption}
\DeclareCaptionFont{tiny}{\tiny}
\usepackage{epstopdf}
\usepackage{amsmath}
\usepackage{amssymb}
\usepackage{multirow}

\title{Diagnosing State-Of-The-Art Object Proposal Methods}

\addauthor{Hongyuan Zhu}{zhuh@i2r.a-star.edu.sg}{1}
\addauthor{Shijian Lu}{slu@i2r.a-star.edu.sg}{1}
\addauthor{Jianfei Cai}{ASJFCAI@ntu.edu.sg}{2}
\addauthor{Guangqing Lee}{134174B@mymail.nyp.edu.sg}{3}
\addinstitution{
Institute for Infocomm Research,\\
A*Star,\\
Singapore
}
\addinstitution{
Nanyang Technological University,\\
Singapore
}
\addinstitution{
	Nanyang Polytechnic,\\
	Singapore
}

\runninghead{H.ZHU, S.LU, J.CAI, G. LEE}{~}

\begin{document}

\maketitle

\begin{abstract}
Object proposal has become a popular paradigm to replace exhaustive sliding 
window search in current top-performing methods in PASCAL VOC and ImageNet. 
Recently, Hosang et al.~\cite{HosangBS14} conduct the first unified study of existing 
methods' in terms of various image-level
degradations. On the other hand, the vital question "what object-level characteristics really affect existing methods' 
performance?" is not yet answered. Inspired by Hoiem et al.'s work in categorical object detection~\cite{Hoiem12}, 
this paper conducts the 
first meta-analysis of various object-level characteristics' impact on state-of-the-art object proposal methods. Specifically, we examine the effects 
of object size, aspect ratio, iconic view, color contrast, shape regularity and texture.
We also analyse existing methods' localization accuracy and latency for various PASCAL VOC object classes. Our study 
reveals the limitations of existing methods in terms of non-iconic view, small object size, low color contrast, shape regularity etc. Based on our observations, lessons are also learned and shared with respect to the selection of existing object proposal technologies as well as the design of the future ones. 
\end{abstract}
\vspace{-20pt}
\section{Introduction}
\label{sec:intro}
Recent top performing methods in PASCAL VOC~\cite{PASCAL10} and ImageNet~\cite{ImageNet} make use of object proposal to replace exhaustive 
window search~\cite{RCNN14, SPH14, ErhanSTA14, CinbisVS13, WangYZL13}.  Object proposal's effectiveness is rooted in the assumption that there are general cues 
to differentiate objects from the background. Since the very first work by Alexe et al.~\cite{Objectness10}, many object proposal methods have been proposed~\cite{Endres10, CPMC10,Rahtu11, SelectiveSearch11, RP13, Bing14,MCG14, EdgeBox14, Rant14, GOP14, RIGOR14} and tested on various large scale datasets~\cite{PASCAL10, COCO, ImageNet}, and their overall detection rates versus different thresholds or window number have also been reported. 
Yet such partial performance summaries 
give us little idea of a method's strengths and weaknesses for further improvement, and users are still facing difficulties in choosing methods for their applications.
Therefore, more detailed analysis of existing state-of-the-arts is critical for future research and applications. 

There are considerable works in categorical object detection~\cite{WuN05, HoiemRW07, WangHY09, DollarWSP09, DivvalaHHEH09} 
which study the impact of different 
object-level properties (such as occlusion, 
aspect ratio, viewpoint change) to the performance of categorical object detectors. 
Others have investigated dataset design~\cite{PintoCD08, TorralbaE11, PASCAL10} 
and impact of the amount of training data~\cite{ZhuVRF12}. 
Our work is greatly inspired by the recent research by Hoeim et al.~\cite{Hoiem12} 
and Russakovsky et al.~\cite{ImageNet} in categorical object detection and Hosang et al.~\cite{HosangBS14} in object proposal.
Hoiem et al.'s work~\cite{Hoiem12} provides depth analysis of existing categorical object detectors in PASCAL VOC in terms of object size, aspect ratio, parts and viewpoints, yet they provided annotations for a limited object classes. 
Russakovsky et al.~\cite{ImageNet} provide further depth analysis of the state-of-the-art categorical object detectors in ILSVRC in terms of various object attributes such as color, shape and texture, though such annotations are not publicly available. Hosang et al.~\cite{HosangBS14} conduct the first unified and comprehensive comparison of existing object proposal methods in terms of various image-level degradations, but the more relevant object-level characteristics analysis are missing.

Our contributions can be summarized in three aspects. First, we investigate the influence of object-level characteristics over state-of-the-art object proposal methods for the first time. Although there are some similar works in categorical object detection, few research has been conducted on object proposal side to the best of our knowledge. Second, we introduce the concept of localization latency to evaluate a method's localization efficiency and accuracy. Third, we create a fully annotated PASCAL VOC dataset with various object-level characteristics to facilitate our analysis. The annotations take us nearly one month's time which will be released to facilitate further related research.
\vspace{-20pt}
\section{Evaluation Protocol}
\subsection{PASCAL VOC detection dataset}
\textbf{Dataset Selection and Annotation}: Our experiments are based on PASCAL VOC2007 test set,
which has been widely used in evaluating object proposal methods~\cite{HosangBS14, Objectness10,Bing14,EdgeBox14}.
The test set has around 15000 object instances across 20 object classes. 
Hoiem et al.~\cite{Hoiem12} make annotations for seven object classes with the characteristics such as object size, aspect ratio, occlusion etc.
In our work, we keep existing annotations for object size and aspect ratio, and extend them to all other object classes. In addition, we make annotations for other object properties (color contrast, shape regularity and textureness), as studied in Russakovsky et al.~\cite{ImageNet}. 
The annotations are conducted by two students and finally confirmed by an expert to reduce the labelling ambiguity. 
\vspace{-2pt}

\textbf{Evaluation Criteria}:
A proposed window $B$ is treated
as detected if its Intersection-over-Union (IoU) with a 
ground truth bounding box $\bar{B}$: $IoU(\bar{B}, B) = \frac{area( B~\cap~\bar{B} )}{area(B~\cup~\bar{B} )} $ is above a certain threshold $T$.
The `best instance IoU' (BIoU) measures the maximum IoU between a ground-truth instance $G~$ with a group of window candidates $S$:
\vspace{-10pt} 
\begin{equation}
 BIoU (G, S) = \underset{s \in S}{\max}~IoU(G, s) 
 \label{eqn: biou}
 \vspace{-10pt}
\end{equation}
The windows of all methods are pre-computed by Hosang et al.~\cite{HosangBS14}.
We adopt the commonly used criterion to evaluate proposal quality
by fixing the number of proposals, then the recall
 varies with different IoU thresholds. 
Due to the page limit, we choose the `recall versus IoU' curve for 1000 windows to
better reflect existing methods' performance in terms of localization accuracy and detection rate.  Numbers next to each method in Fig~\ref{fig:iou_recall_1000_man_made}$\sim$~\ref{fig:iou_recall_1000_texture} indicate area under curve and average window number per image, respectively.
We also attach the figures for other window number (100 and 10000) in supplementary 
material for reference. The curves for `recall versus region number' and `area under recall-versus-IoU' are also attached to provide complementary information. 
\vspace{-10pt}
\subsection{Object Proposal Methods}
We choose representative and recent top-performing object proposal methods in PASCAL VOC 2007 dataset as 
our experiment candidates to reflect the recent development. Existing methods can be divided into two major groups,
window based methods and region based methods. 

\textbf{\textit{Window based methods}} generate proposals by ranking each sampled window according
to the likelihood that it contains the object of interest. 

1)~\textbf{Objectness (O)~\cite{Objectness10}} assigns the objectness likelihood by sampling initial windows from salient
object locations and then the cues from color contrast, edge distribution, superpixel straddling are integrated by using Bayes rule. 

2)~\textbf{Rathu2011 (R1)~\cite{Rahtu11}} extends~\cite{Objectness10} by using new objectness cues from superpixel and image gradients, and applies a structured output ranking method to combine them. 

3)~\textbf{BING (B)~\cite{Bing14}} ranks coarse to fine sliding windows by using classifiers trained with the normalized gradient feature. It encloses the closed contour information and can run at 300fps on a desktop.

4)~\textbf{EdgeBox (EB)~\cite{EdgeBox14}} also evaluates the objectness with sliding windows, but the scoring is derived
from the state-of-the-art structured forest contour detector~\cite{SF}. Each window is assigned a score corresponds
to the strength of enclosed contour signal. 
\\\vspace{-10pt}

\textbf{\textit{Region based methods}} generate multiple foreground regions that correspond
to objects. The process starts from single or multiple over-segmentation, then the regions
are grouped according to the multiple cues and finally the candidate regions are ranked either 
according to region hierarchy or by using a learned ranking classifiers etc. A region proposal
can be transformed to a window proposal by enclosing the region with the tightest bounding box. 

1)~\textbf{CPMC (C)~\cite{CPMC10}} applies non-overlapped seeds to train appearance models for foreground and background and solves
graph cuts to generate foreground proposals. Then the proposals are ranked according to various cues. Endres et al.~\cite{Endres10} applies a similar pipeline but using a seeding map from occlusion boundary and a different ranking model.

2)~\textbf{MCG (M)~\cite{MCG14}} makes use of a novel multi-scale eigenvector solver to quickly globalize the object contours. Then the regions are merged to form candidates according to the contour similarities in different scales. Further candidates are generated by merging up to four regions, and are ranked according to various cues similar to CPMC.
 
3)~\textbf{SelectiveSearch (SS)~\cite{SelectiveSearch11}} also starts from superpixels, but it applies a diversified approach to form object regions by merging them with a set of hand-crafted features and similarity functions. 

4)~\textbf{RandomizedPrim (RP)~\cite{RP13}} follows a similar approach to SelectiveSearch, while the weights between superpixels are learned and the merging process is randomized. 

5)~\textbf{Geodesic (G)~\cite{Geo14}} assigns seeds by using heuristics or learned classifiers for subsequent geodesic distance transform. The level sets of each geodesic transform define and rank the foregrounds. 

6)~\textbf{RIGOR (RI)~\cite{RIGOR14}} is similar to CPMC, but the contour detector is replaced with the fast structure forest~\cite{SF} and the algorithm is speeded up by reusing the inference. 
\\\vspace{-10pt}

\textbf{\textit{Baselines}} are also considered in our work to provide reference points to study the correlation between the existing methods and their components. We consider two baselines from Hosang et al.~\cite{HosangBS14} in this work. 1)~\textbf{Uniform (U)} baseline generates proposals by uniformly sampling the object center, square root area and aspect ratio. 2)~\textbf{Superpixel (SP)} of~\cite{FelzenszwalbH04} is adopted, as four methods in this work apply its superpixel. This baseline over-segments an image into regions where each region is treated as a candidate. 
\vspace{-15pt}
\section{Profiling Localization Accuracy and Latency}
\label{sec:profiling}
\begin{table}[htb]\tiny
	\begin{tabular}{p{5pt}p{5pt}p{5pt}p{5pt}p{5pt}p{5pt}p{5pt}p{5pt}p{5pt}p{5pt}p{5pt}p{5pt}p{5pt}p{5pt}p{5pt}p{5pt}p{5pt}p{5pt}p{5pt}p{5pt}p{5pt}p{5pt}}
		\hline
		Method & Person & Bird & Cat & Cow & Dog & Horse & Sheep & Plane & Bike & Boat & Bus & Car & MBike & Train & Bottle & Chair & Table & Plant & Sofa & TV & mIoU\\
		\hline
		Bing & 0.58 & 0.57 & 0.70 & 0.58 & 0.67 & 0.63 & 0.57 & 0.61 & 0.61 & 0.53 & 0.62 & 0.56 & 0.62 & 0.67 & 0.49 & 0.55 & 0.65 & 0.56 & 0.68 & 0.59 & 0.60\\
		EB & \textbf{0.65} & \textbf{0.69} & \textbf{0.80} & \textbf{0.72} & \textbf{0.80} & \textbf{0.77} & \textbf{0.70} & \textbf{0.74} & \textbf{0.74} & \textbf{\color{red}0.64} & \textbf{0.78} & \textbf{0.66} & \textbf{0.74} & \textbf{0.78} & \textbf{0.52} & \textbf{0.63} & \textbf{0.74} & \textbf{0.63} & 0.77 & \textbf{0.76} & \textbf{0.71}\\
		OBJ & 0.57 & 0.57 & 0.72 & 0.59 & 0.69 & 0.67 & 0.56 & 0.64 & 0.63 & 0.54 & 0.68 & 0.57 & 0.63 & 0.70 & 0.45 & 0.53 & 0.70 & 0.54 & 0.72 & 0.60 & 0.61\\
		Rathu& 0.57 & 0.57 & 0.79 & 0.61 & 0.77 & 0.73 & 0.59 & 0.69 & 0.67 & 0.53 & 0.73 & 0.57 & 0.69 & 0.77 & 0.38 & 0.50 & \textbf{0.74} & 0.52 & \textbf{0.79} & 0.68 & 0.64\\
		\hline
		Mean & 0.59 &  0.60 &  0.75 & 0.63 & 0.73 & 0.70 & 0.60 & 0.67 & 0.66 & 0.56 & 0.70 & 0.59 & 0.67 &    0.73 & 0.46 &  0.55 &  0.71 & 0.56 & 0.74 & 0.66 & 0.64 \\
		\hline
		CPMC & 0.62 & 0.65 & \textbf{\color{red}0.87} & 0.70 & \textbf{\color{red}0.85} & 0.76 & 0.66 & 0.72 & 0.68 & 0.55 & 0.77 & 0.64 & 0.72 & 0.80 & 0.46 & 0.61 & 0.76 & 0.59 & 0.85 & 0.74 & 0.70\\
        GOP  & 0.66 & 0.65 & 0.86 & 0.72 & 0.82 & 0.77 & 0.68 & 0.69 & 0.72 & 0.59 & 0.80 & 0.69 & 0.74 & 0.81 & 0.52 & 0.66 & 0.78 & 0.64 & 0.83 & 0.76 & 0.72\\
		MCG & \textbf{\color{red}0.71} & \textbf{\color{red}0.70} & \textbf{\color{red}0.87} & \textbf{\color{red}0.77} & \textbf{\color{red}0.85} & \textbf{\color{red}0.79} & \textbf{\color{red}0.73} & 0.76 & \textbf{\color{red}0.75} & \textbf{\color{red}0.64} & \textbf{\color{red}0.82} & \textbf{\color{red}0.72} & \textbf{\color{red}0.77} & \textbf{\color{red}0.83} & \textbf{\color{red}0.58} & \textbf{\color{red}0.69} & 0.78 & \textbf{\color{red}0.65} & \textbf{\color{red}0.87} & \textbf{\color{red}0.80} & \textbf{\color{red}0.75}\\
		RP & 0.63 & 0.64 & 0.85 & 0.68 & 0.82 & 0.73 & 0.66 & 0.77 & 0.71 & 0.59 & 0.77 & 0.65 & 0.73 & 0.78 & 0.49 & 0.65 & 0.79 & 0.61 & 0.86 & 0.75 & 0.71\\
        RIGOR & 0.61 & 0.64 & 0.87 & 0.69 & 0.82 & 0.75 & 0.64 & 0.69 & 0.70 & 0.57 & 0.76 & 0.65 &  0.72 & 0.78 & 0.48 & 0.62 & 0.74 &    0.59 &  0.84 & 0.72 & 0.69\\
		SS & 0.67 & 0.69 & \textbf{\color{red}0.87} & 0.72 & \textbf{\color{red}0.85} & 0.77 & 0.68 & \textbf{\color{red}0.79} & \textbf{\color{red}0.75} & 0.63 & 0.79 & 0.68 & 0.76 & 0.82 & 0.53 & 0.67 & \textbf{\color{red}0.82} & 0.64 & \textbf{\color{red}0.87} & 0.77 & 0.74\\
		\hline
		Mean &  0.65 &  0.66 &  0.86 &  0.71 & 0.83 &  0.76 & 0.68 & 0.74 & 0.72 &  0.60 & 0.78 &  0.67 &  0.74 & 0.80 & 0.5083 &   0.65 &   0.78 &  0.62 &  0.85 & 0.76 & 0.72\\
		\hline
	\end{tabular}	
	\\
	\\	
	\caption{\small Different methods' localization accuracy for each PASCAL VOC class, which is defined in Sec~\ref{sec:profiling}. The methods on the upper parts of the rows are the window based methods, the methods following rows are for the region based methods. Bold face numerics are the best results for each methodology. While, the bold face numerics in red color demonstrates the bests results for all methods. }
	\label{tab:mean_best_iou}
\end{table}

To motivate our object level analysis, we need to answer the question \textit{`Are state-of-the-arts really good at all kinds of objects?'}. Therefore, we juxtapose each PASCAL VOC class' localization accuracy in Table~\ref{tab:mean_best_iou}. The~\textbf{localization accuracy} is measured by taking average of the 'mean best instance IoU' for different window number $L\in\{100, 1000,  10000\}$, where `mean best instance IoU' measures the mean BIoU (Eq.~\ref{eqn: biou}) between all ground-truth instances and a group of window candidates $S$, given $|S| \in L$. 
As demonstrated in Table~\ref{tab:mean_best_iou}, different methods' localization accuracy varies with classes, which gives hints that object proposals are not as `object agnostic' as the original assumption. The region based methods have higher localization accuracy than window based methods. MCG and SelectiveSearch are the top performing region based methods, though window based EdgeBox shows comparable performance. The localization accuracy for region based methods are similar. One potential explanation is that all region based methods follow similar pipeline by grouping superpixels 
with either learned or hand-crafted edge measures. 
\vspace{-2pt}
\begin{table}[htb]\tiny
\begin{tabular}{p{5pt}p{5pt}p{5pt}p{5pt}p{5pt}p{5pt}p{5pt}p{5pt}p{5pt}p{5pt}p{5pt}p{5pt}p{5pt}p{5pt}p{5pt}p{5pt}p{5pt}p{5pt}p{5pt}p{5pt}p{5pt}p{5pt}p{5pt}}
\hline
Method & Person & Bird & Cat & Cow & Dog & Horse & Sheep & Plane & Bike & Boat & Bus & Car & MBike & Train & Bottle & Chair & Table & Plant & Sofa & TV & Mean\\
\hline
Bing & \textbf{\color{red}18.2} & 15.1 &\textbf{\color{blue}13.4} & 14.3 & 14.2 & 14.1 & 14.2 & 14.1 & 14.3 & 14.7 &\textbf{\color{blue}13.6} &\textbf{\color{red}16.6} & 14.1 &\textbf{\color{blue}13.6} &\textbf{\color{red}15.6} &\textbf{\color{red}16.4} &\textbf{\color{blue}13.7} &\textbf{\color{red}15.2} &\textbf{\color{blue}13.9} & 14.3 & 14.7\\  
EB & \textbf{\color{red}18} & 14.6 &\textbf{\color{blue}13.0} & 13.6 & 13.6 &\textbf{\color{blue}13.4} & 13.7 & 13.5 & 13.8 & 14.3 &\textbf{\color{blue}12.7} &\textbf{\color{red}16.1} &13.6 &\textbf{\color{blue}13.1} &\textbf{\color{red}15.4} &\textbf{\color{red}16.2} & 13.5 &\textbf{\color{red}15} & 13.5 &\textbf{\color{blue}13.4} & \textbf{14.2} \\
Obj & \textbf{\color{red}18.3} & 15.1 &\textbf{\color{blue}13.4} & 14.2 & 14.2 & 14.0 & 14.3 & 13.9 & 14.3 & 14.7 &\textbf{\color{blue}13.3} &\textbf{\color{red}16.5} & 14.2 &\textbf{\color{blue}13.3} &\textbf{\color{red}15.7} &\textbf{\color{red}16.6} &\textbf{\color{blue}13.5} &\textbf{\color{red}15.3} &\textbf{\color{blue}13.6} & 14.3 & 14.6\\ 
Rathu & \textbf{\color{red}18.5} &15.3 &\textbf{\color{blue}14.0} & 14.4 & 14.7 & 14.5 & 14.4 &\textbf{\color{blue}14.1} & 14.5 & 14.8 &\textbf{\color{blue}13.6} &\textbf{\color{red}16.6} &14.4 &\textbf{\color{blue}13.8} &\textbf{\color{red}15.8} &\textbf{\color{red}16.7} &\textbf{\color{blue}13.8} &\textbf{\color{red}15.5} & 14.1 & 14.3 & 14.9\\
\hline
Mean  & 18.3  & 15.0 &   13.4 &  14.1 & 14.2 &  14.0 & 14.1 & 13.9 & 14.2 & 14.6 & 13.3 &  16.5 & 14.1 &  13.4 & 15.6 & 16.5 & 13.6 &  15.2 & 13.8 & 14.1 & 14.6\\
\hline
CPMC & \textbf{\color{red}18.3} & 14.9 &\textbf{\color{blue}13.2} & 13.9 & 13.9 & 14.0 & 14.0	& 13.8 & 14.4 & 14.7 &\textbf{\color{blue}13.4} &\textbf{\color{red}16.4} & 14.1 &\textbf{\color{blue}13.4} &\textbf{\color{red}15.7} &\textbf{\color{red}16.4} &\textbf{\color{blue}13.7} &\textbf{\color{red}15.3} &\textbf{\color{blue}13.6} & 14 & 14.6\\
GOP & \textbf{\color{red}18.4} & 15.1 & \textbf{\color{blue}13.9} & \textbf{\color{blue}14.0}	& 14.5 & 14.4 & \textbf{\color{blue}14.0} & 14.1 & 14.5 & 14.7 & \textbf{\color{blue}13.6} & \textbf{\color{red}16.4} & 14.5 & \textbf{\color{blue}14.0} & \textbf{\color{red}15.6} & \textbf{\color{red}16.4} & \textbf{\color{blue}14.0} & \textbf{\color{red}15.3} & 14.1 & \textbf{\color{blue}14.0} & 14.8\\  
MCG & \textbf{\color{red}18.2} & 14.9 &\textbf{\color{blue}13.3} &\textbf{\color{blue}13.8} & 14.1 & 14.1 &\textbf{\color{blue}13.8} & \textbf{\color{blue}13.8} & 14.3 & 14.6 &\textbf{\color{blue}13.3} &\textbf{\color{red}16.3} & 14.1 &\textbf{\color{blue}13.6} &\textbf{\color{red}15.5} &\textbf{\color{red}16.3} & \textbf{\color{blue}13.9} &\textbf{\color{red}15.2} & 13.8 & \textbf{\color{blue}13.9} & \textbf{14.5}\\
RP & \textbf{\color{red}18.5} & 15.2 &\textbf{\color{blue}14.1} & 14.3 & 14.8 & 14.6 & 14.3 &\textbf{\color{blue}14.0} & 14.6 & 14.7 &\textbf{\color{blue}13.9} &\textbf{\color{red}16.7} & 14.5 &\textbf{\color{blue}14.1} &\textbf{\color{red}15.7} &\textbf{\color{red}16.5} &\textbf{\color{blue}14.1} &\textbf{\color{red}15.4} & 14.2 & 14.3 & 14.9 \\ 
RIGOR & \textbf{\color{red}18.3} & 14.9 & \textbf{\color{blue}13.1} & 14.0 & 13.9 & 14.0 & 14.0 & 13.7 & 14.2 & 14.6 & \textbf{\color{blue}13.3} &\textbf{\color{blue}16.3} & 14.0 & \textbf{\color{blue}13.3} &\textbf{\color{red}15.6} &\textbf{\color{red}16.3} &\textbf{\color{blue}13.6} & \textbf{\color{red}15.2} &\textbf{\color{blue}13.3} & 14.0 & \textbf{14.5}\\  
SS & \textbf{\color{red}18.4} & 15.1 &\textbf{\color{blue}13.8} & 14.2 & 14.4 & 14.4 & 14.2 &\textbf{\color{blue}13.7} & 14.4 & 14.7 &\textbf{\color{blue}13.6} &\textbf{\color{red}16.6} & 14.3 &\textbf{\color{blue}13.8} &\textbf{\color{red}15.7} &\textbf{\color{red}16.5} &\textbf{\color{blue}13.8} &\textbf{\color{red}15.4} & 14 & 14.3 & 14.8\\
\hline 
Mean & 18.3 & 15.0 & 13.6 & 14.0 & 14.3 & 14.2 & 14.0 & 13.9 &  14.4 & 14.7 & 13.5 & 16.5 &  14.2 &  13.7 & 15.6 & 16.4  & 13.9 & 15.3 &  13.8 &  14.1 & 14.7 \\
\hline 	
\end{tabular}
\vspace{5pt}
\caption{Different methods' localization latency for each PASCAL VOC class, which is defined in Eq.~\ref{eqn: LD}. The methods in the top part of rows are the window based approach, the methods in the below rows are for the region based approach. The blue numerics indicate the top 5 lowest latency classes for each method, the red ones indicate the top 5 highest latency classes.}
\label{tab: LD}
\end{table}

A good object proposal method should not only produce candidates with high accuracy, but also use as less windows as possible. To summarize a method's performance in terms of the accuracy and window number, we propose the \textbf{localization latency} metric as inspired by the clutterness measure in~\cite{Russakovsky13}, which measures
the average number of windows required to localize all object instances.
Moreover, different object classes intuitively should have different levels of localization latency. For example, some classes (e.g. sheep or cow) often stay in a clean background, and are easy to be localized with a few windows. On the other hand, when an object is located in a cluttered indoor, more windows may be needed to localize the object. By studying the localization latency, we can have some initial idea of what object appearance properties may affect the performance of existing methods. 

To quantify a method $I$'s localization latency, we sample $K~(K\in\{100, 1000, 10000\})$ windows $W_1^I, W_2^I,..., W_K^{I}$ in returned descending order. Let $OBJ(m, T, K)$ be the total number of sampled windows to localize all instances of the target category in image $m$ under varying threshold $T\in \{0.5, 0.7, 0.85\}$ and $K$ (i.e. $OBJ(m, T, K) = \sum_i min\{k \in K: IOU(W_k^m, B_i^m) > T \}$). If using $k$ windows still can't localize the object, $k$ is set to $K+1$. $M$ is the number of images contain the instances. The localization latency is defined as follows:
\begin{equation}
 LocalizationLatency =  \frac{1}{|T|}\sum_{T}\log_2(\frac{1}{M}\sum_{K}\sum_m OBJ(m,T, K))
 \label{eqn: LD}
\end{equation}

Table~\ref{tab: LD} juxtaposes the class localization latency of the selected methods. The higher the measure, the more windows are needed to localize the object. The top 5 most difficult classes are `Person', `Car', `Chair', `Plant' and `TV', where the objects contain strong intra-variation, thin structures and often from images with background clutter. On the other hand, the top 5 easiest classes are `Cats', `Bus', `Train', `Table' and `Plane', which tend to have large homogeneous foreground or background or are highly contrasted. It can therefore be concluded that some object properties can indeed influence the performance of existing proposal methods. Moreover, by inspecting from the table, the window based methods have lower localization latency than region based methods. 
Our interpretation for such result is that window based approach inherently assumes that objects are spatially compact. As a result, the window sampled by using carefully hand-crafted parameters can capture more integrated objects in the combinatorial space, which are difficult to be detected by using bottom-up region based approach. Finally, the measures also reflect complementariness among different methods, which is difficult to be revealed by overall IoU metric.
\vspace{-15pt}
\section{Impact of Object Characteristics}
So far, we have studied existing methods' localization accuracy and latency, and
can conclude that object appearance indeed influences the performance of different proposal methods. 
We look further into what object properties can bring such influences. First, we study the impact of natural and man-made objects.
Then, we examine the impact of additional properties: iconic view, object size, aspect ratio, color contrast, shape regularity and textureness.

\begin{figure}[h]
	\begin{minipage}[b]{.45\linewidth}	
		\centering
		\setlength{\tabcolsep}{0.5pt}
		\renewcommand{\arraystretch}{0.5}
		\begin{tabular}{cc}
			\includegraphics[width=.5\linewidth]{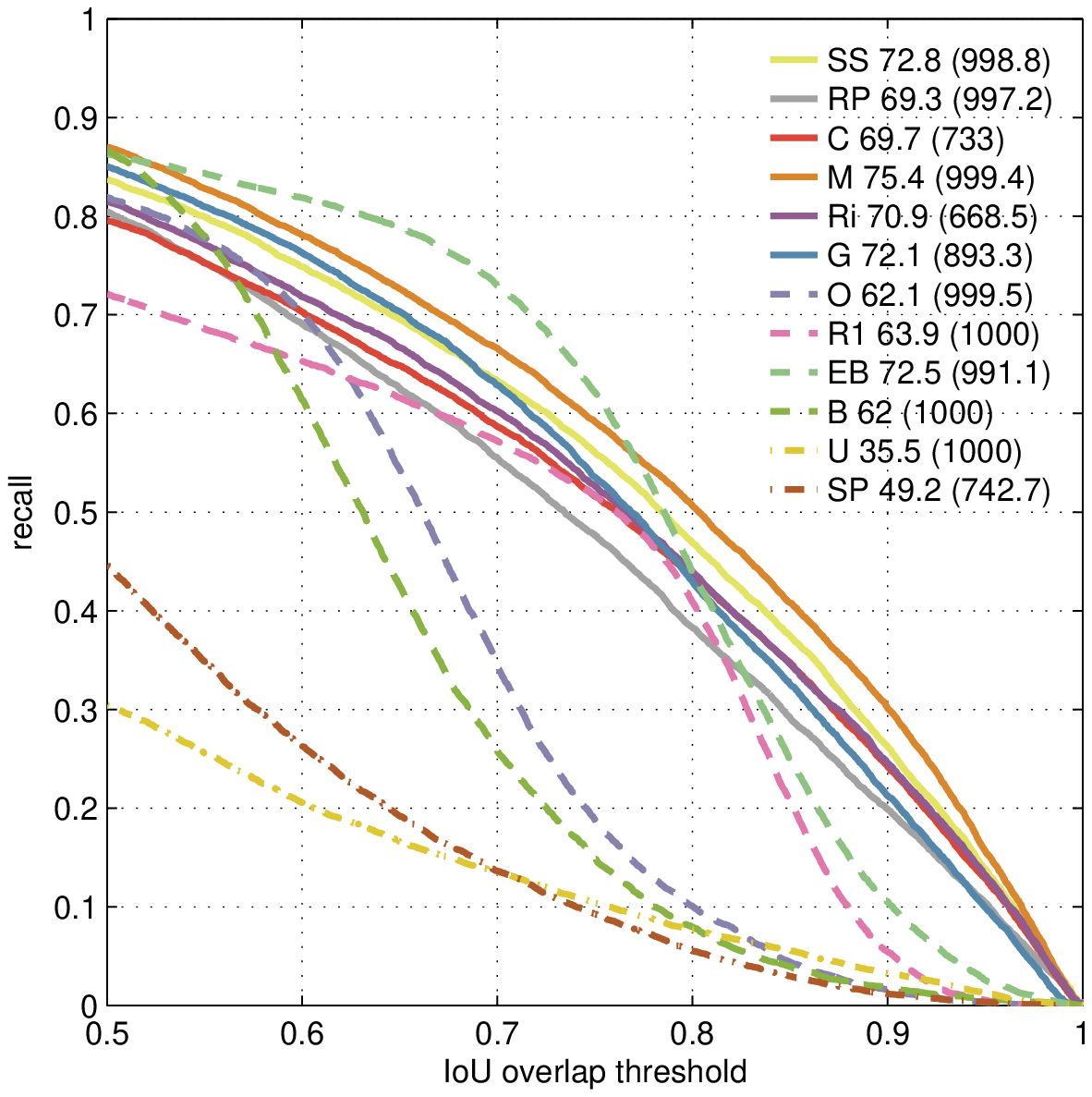}&
			\includegraphics[width=.5\linewidth]{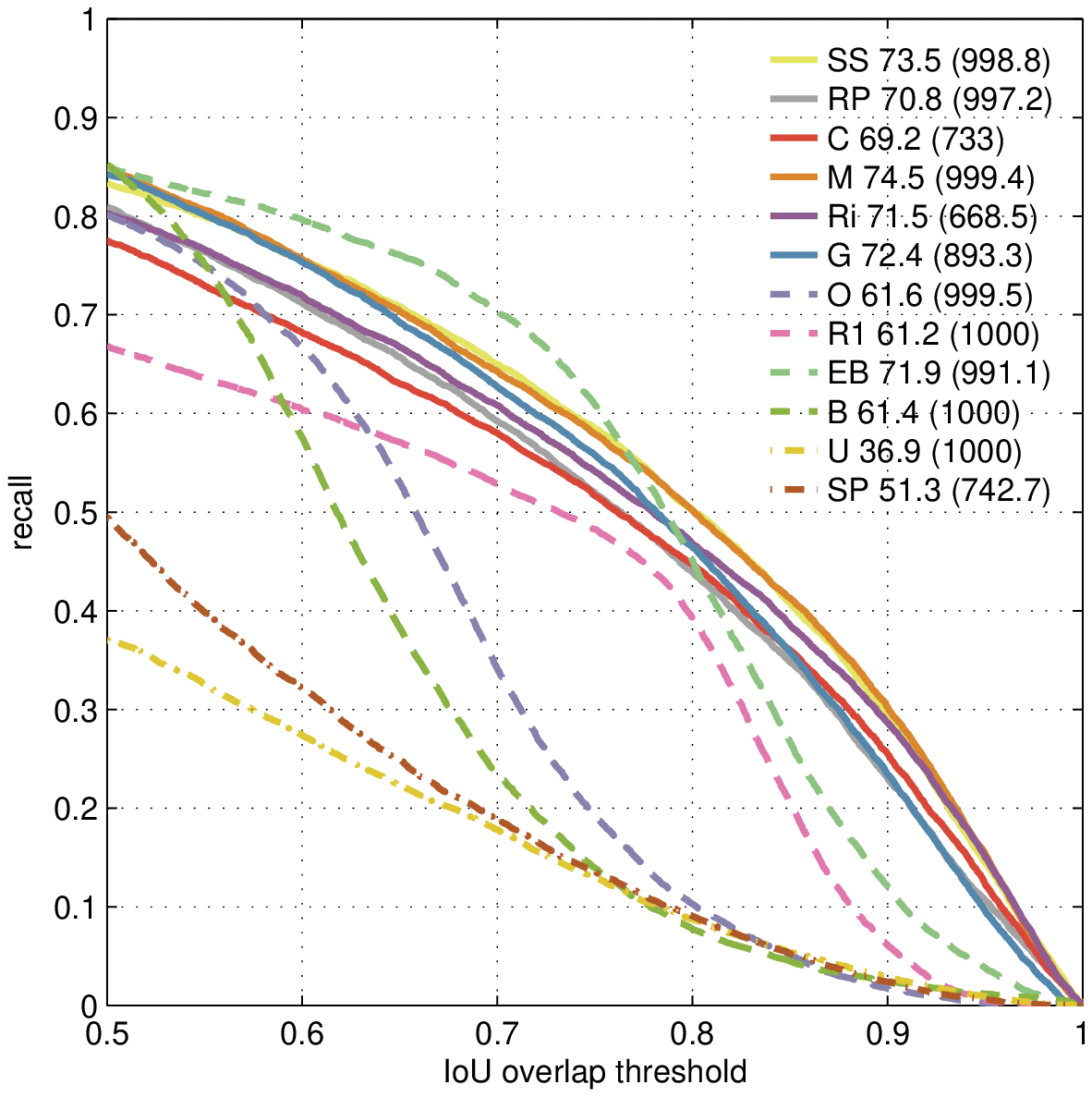}\\
			\tiny(a)&\tiny(b)
		\end{tabular}
		\vspace{3pt}
		\caption{\footnotesize Recall versus IoU threshold curves for \textit{`Natural'} (a) and \textit{`Man-Made'} (b). Dashed lines are for window based methods and solid lines are for region based methods. Dashed lines with dots are for the baseline methods.}
		\label{fig:iou_recall_1000_man_made}
	\end{minipage}
	\hspace{20pt}
	\begin{minipage}[b]{.45\linewidth}
		\centering	
		\setlength{\tabcolsep}{0.5pt}
		\renewcommand{\arraystretch}{0.5}
		\begin{tabular}{cc}
			\includegraphics[width=.5\linewidth]{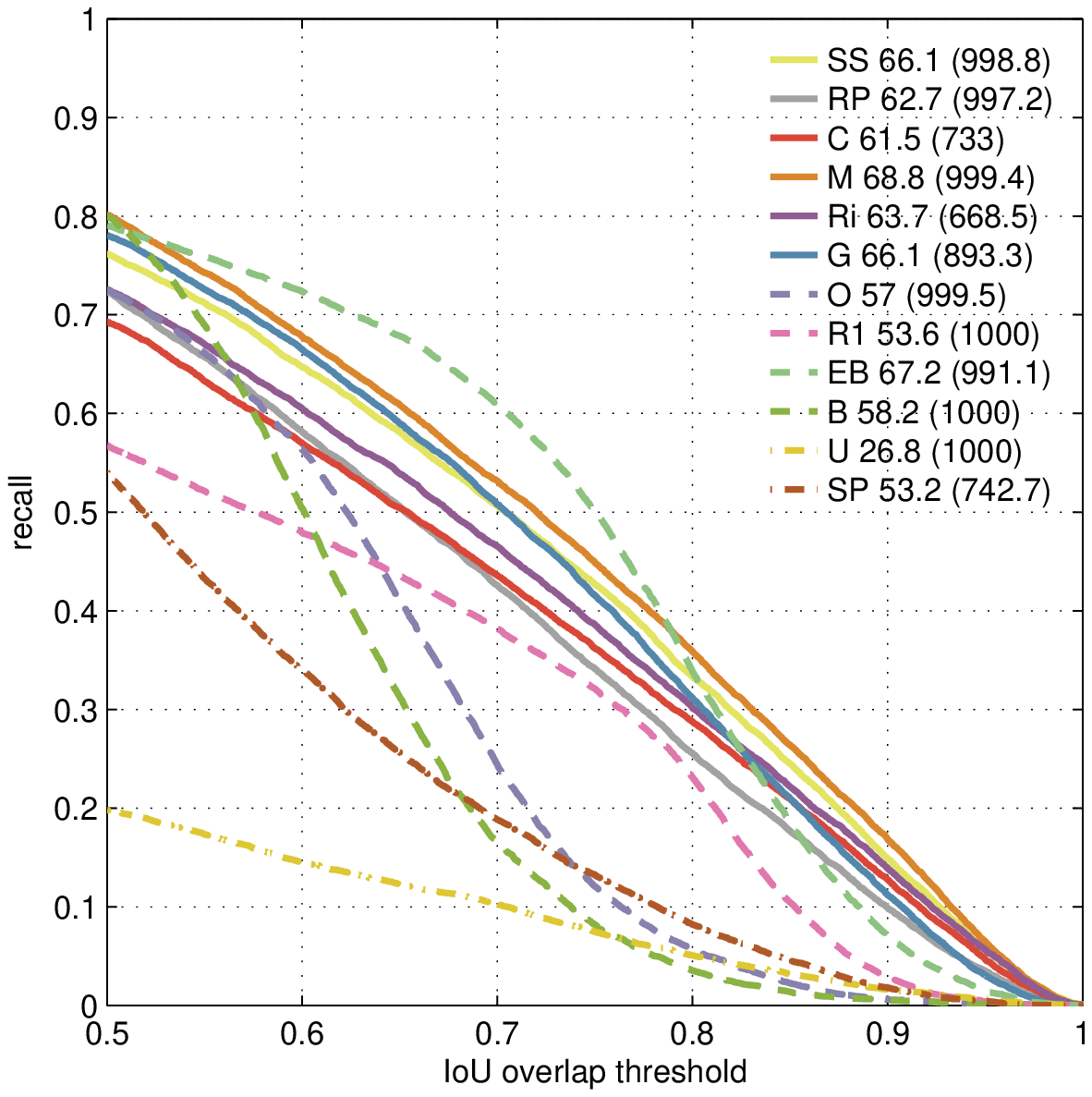}&
			\includegraphics[width=.5\linewidth]{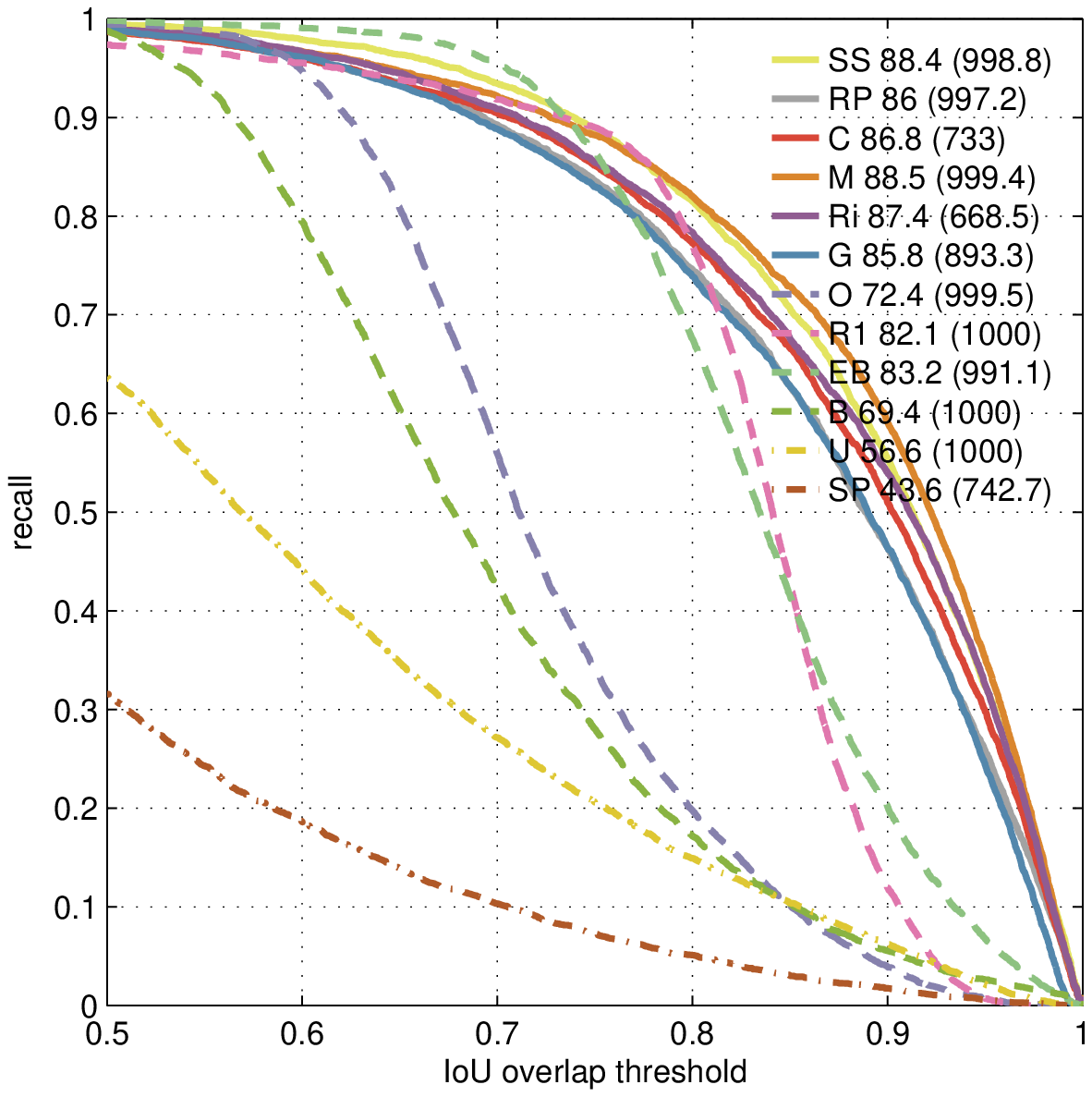}\\
			\tiny(a)&\tiny(b)
		\end{tabular}
		\vspace{3pt}
		\caption{\footnotesize Recall versus IoU threshold curves for \textit{`Non-iconic'} (a) and \textit{`Iconic'} (b) objects. Dashed lines are for window based methods and solid lines are for region based methods. Dashed lines with dots are for the baseline methods.}
		\label{fig:iou_recall_1000_iconic}
	\end{minipage}
	\vspace{-10pt}
\end{figure}
\textbf{Natural vs Man-made} classifies objects as `natural' (e.g. horse, person, cow et al) or 
`man-made' (e.g. motorbike, tv monitor et al). The `recall versus IoU threshold' curves for \textit{`Natural'} and \textit{`Man-Made'} objects are shown in Figure ~\ref{fig:iou_recall_1000_man_made}. The
recall rate for natural objects is slightly higher than man-made
ones, though the difference is almost negligible. This
suggests existing objectness rules are general to man-made and nature objects. Region based methods maintain 
a smooth recall rate transition, while the window based EdgeBox achieves the highest recall when it is tuned for $IoU = 0.7$. 

\textbf{Iconic Objects} refers to the objects in canonical perspective centred in the image with certain size, and vice versa. Many commercial photos contain such `iconic objects'. On the other hand, for images such as street or indoor scenes, objects appear in various locations, scale and aspect ratio. Therefore, we are interested in studying whether existing methods would favour such `iconic objects'. 
To decide whether an object is iconic or not, we
estimate a multivariate Gaussian distribution for the bounding box parameters e.g. centre position, square root area, and log aspect ratio from the PASCAL VOC's training set as Hosang et al.~\cite{HosangBS14}. Then we sample bounding boxes from this distribution. An iconic object is decided 
by checking if there is an sampled bounding box that has an $IoU \geq 0.6$
with the ground truth. Non-iconic objects take up nearly $68\%$ of the objects. The ``recall versus IoU threshold'' curves for \textit{`Non-iconic'} and \textit{`Iconic'} objects are shown in Figure ~\ref{fig:iou_recall_1000_iconic}. To our surprise, existing methods strongly favour iconic objects. One conjecture is that some methods already incorporate the prior knowledge for iconic object by applying classifiers with location features (e.g. MCG, CPMC, RIGOR) or window sampling technique (e.g. Rathu2011). On the other hand, many methods that don't incorporate such prior (e.g. BING, EdgeBox, Objectness, SelectiveSearch, Geodesic) also exhibit different degrees of performance loss, though such loss is smaller compared with the ones apply the location prior (except MCG).  Such phenomenon suggests the existing methods have limitations in localizing objects in estrange positions and appearances. Our subsequent analysis also supports such hypothesis.
\begin{figure*}[h]
	\begin{minipage}[b]{1\linewidth}	
		\setlength{\tabcolsep}{0.5pt}
		\renewcommand{\arraystretch}{0.5}
		\centering
		\begin{tabular}{ccccc}
			\includegraphics[width=.2\linewidth]{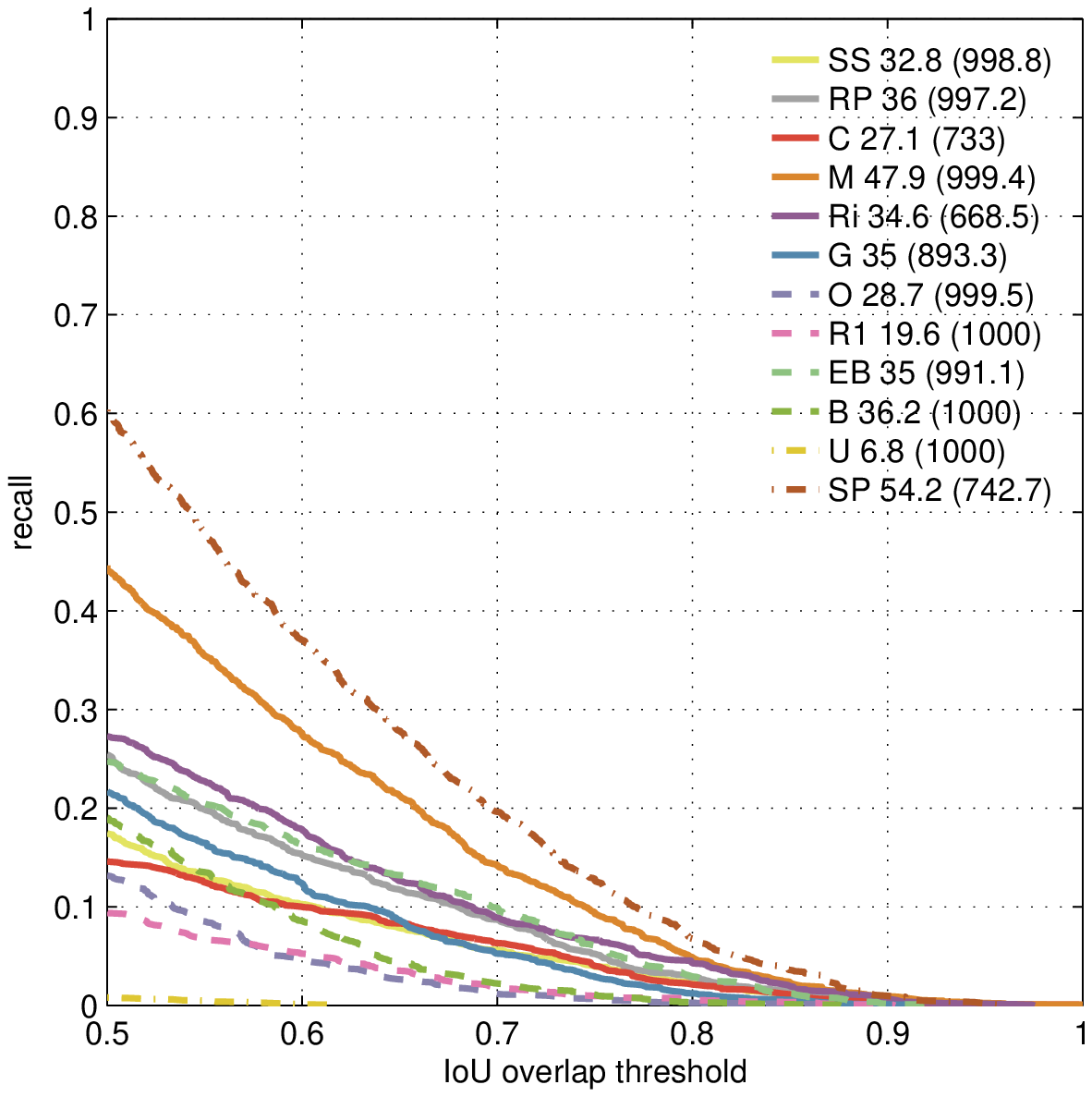}&
			\includegraphics[width=.2\linewidth]{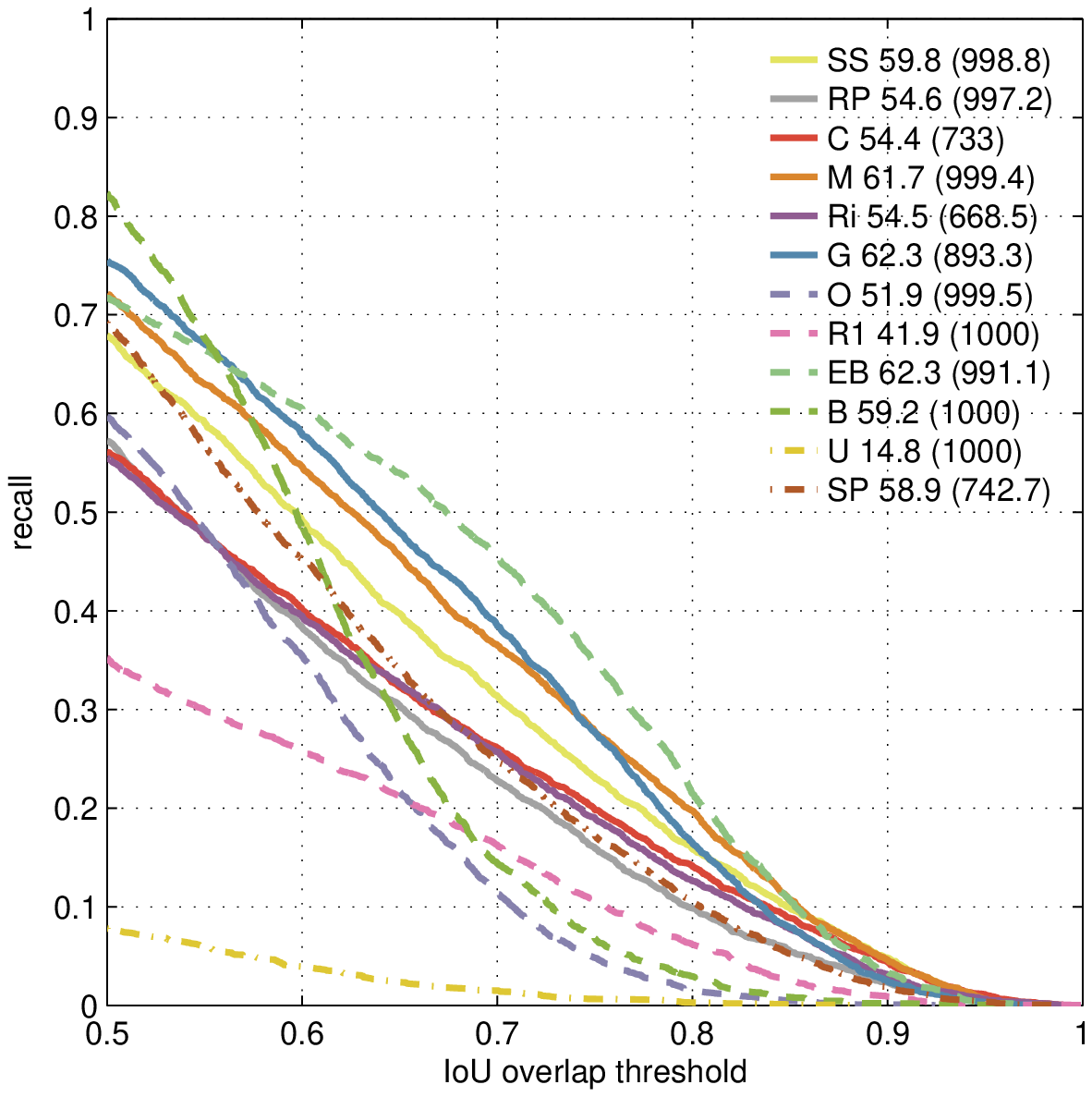}&		\includegraphics[width=.2\linewidth]{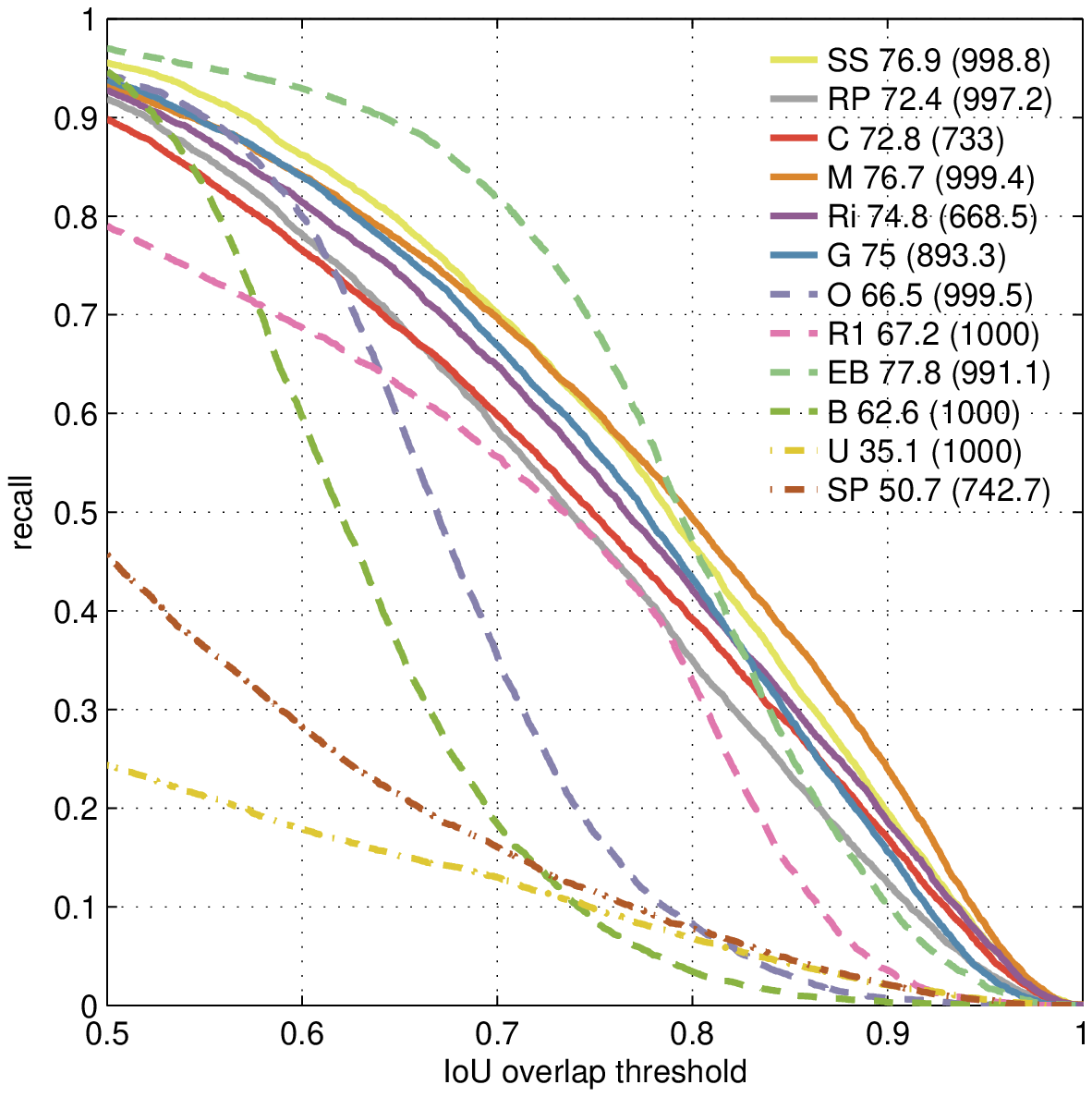}&
			\includegraphics[width=.2\linewidth]{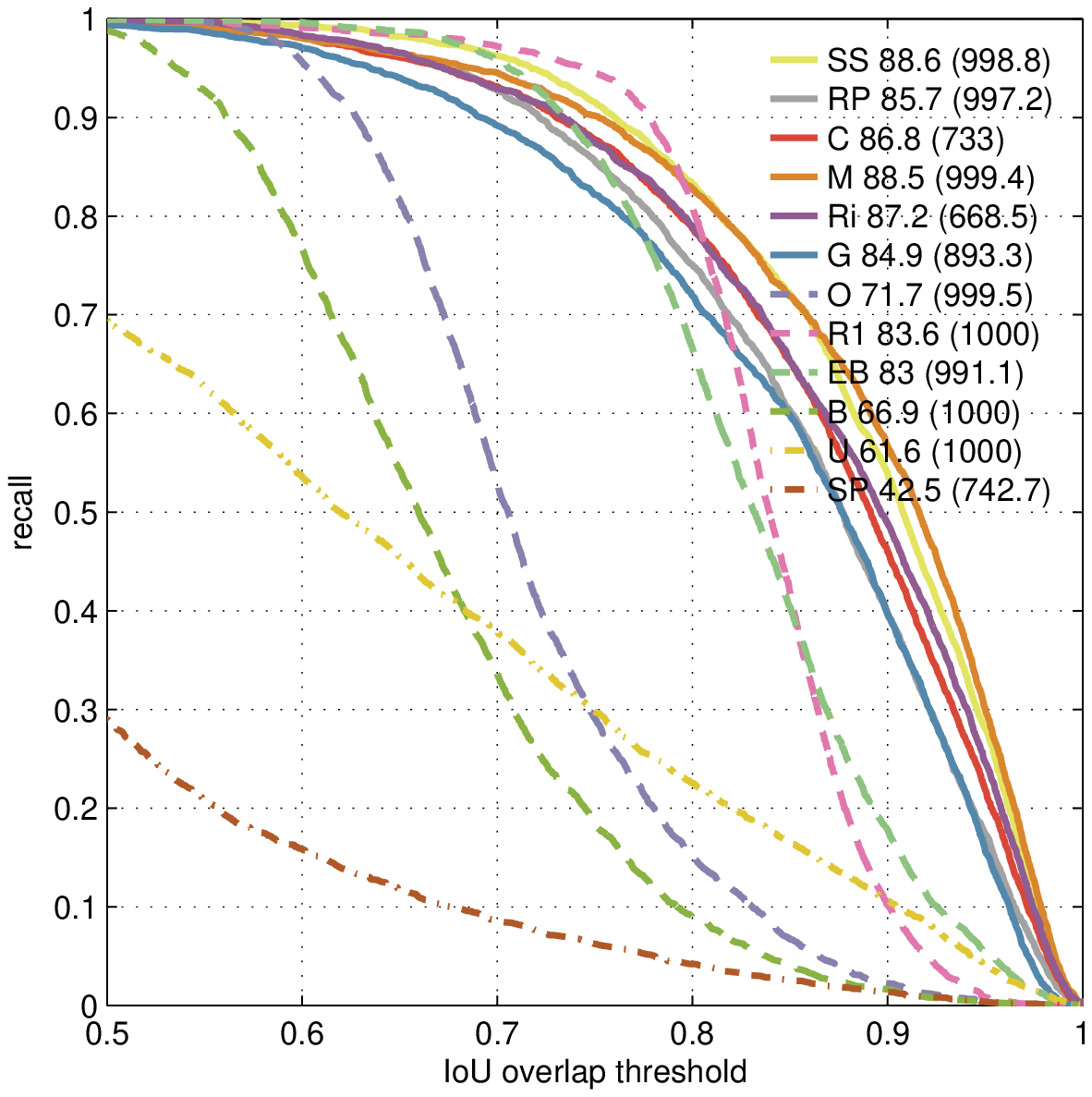}&
			\includegraphics[width=.2\linewidth]{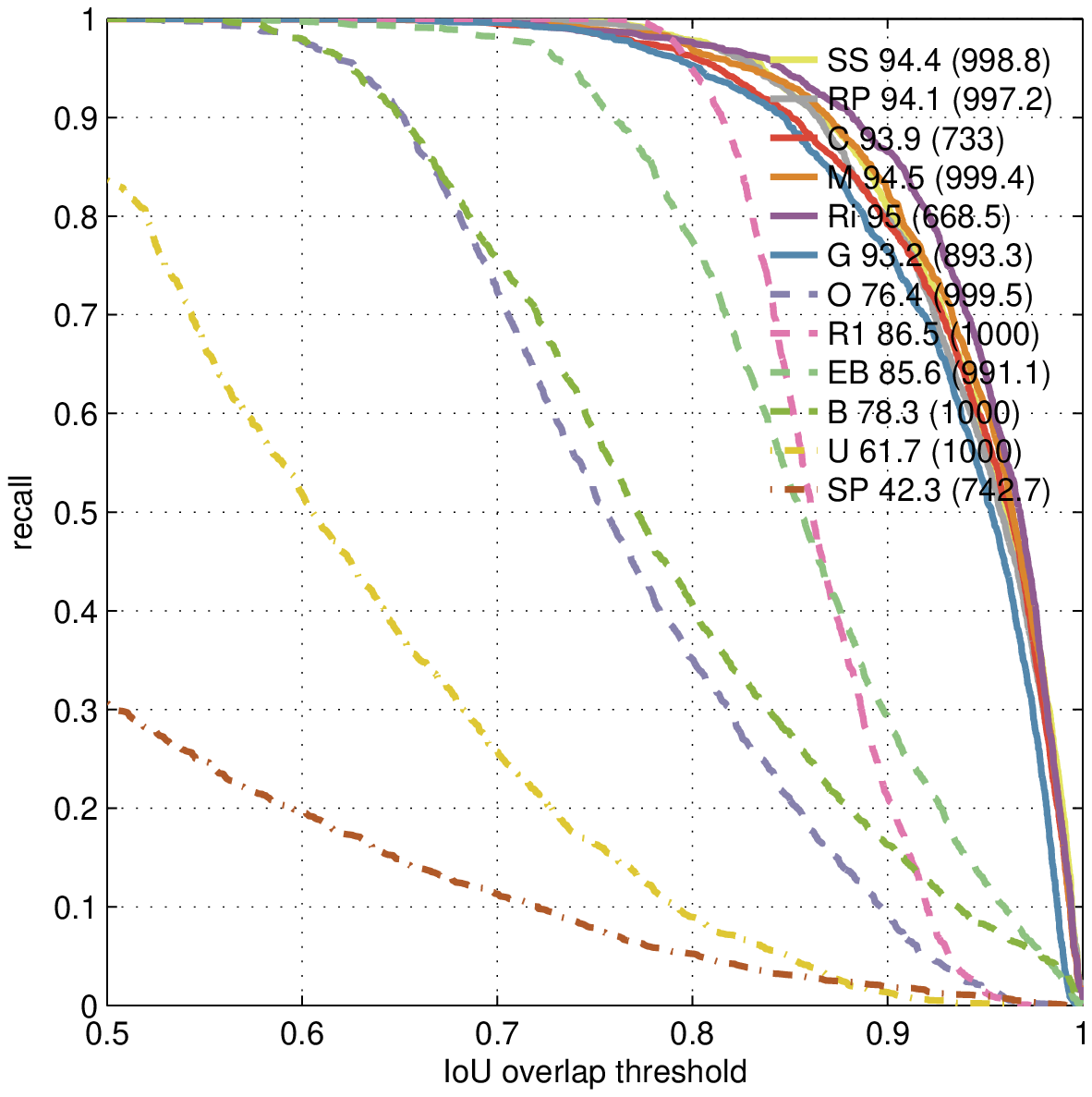}\\
			\tiny(a)&\tiny(b)&\tiny(c)&\tiny(d)&\tiny(e)
		\end{tabular}
		\vspace{3pt}
		\caption{\footnotesize Recall versus IoU threshold curves for different object sizes: \textit{`extra-small'} (a), \textit{`small'} (b), \textit{`medium'} (c), \textit{`large'} (d) and \textit{`extra-large'} (c). Dashed lines are for window based methods and solid lines are for region based methods. Dashed lines with dots are for the baseline methods.}
		\label{fig:iou_recall_1000_object_size}
	\end{minipage}
	\\
	\begin{minipage}[b]{1\linewidth}	
		\centering
		\setlength{\tabcolsep}{0.5pt}
		\renewcommand{\arraystretch}{0.5}
		\begin{tabular}{ccccc}
			\includegraphics[width=.2\linewidth]{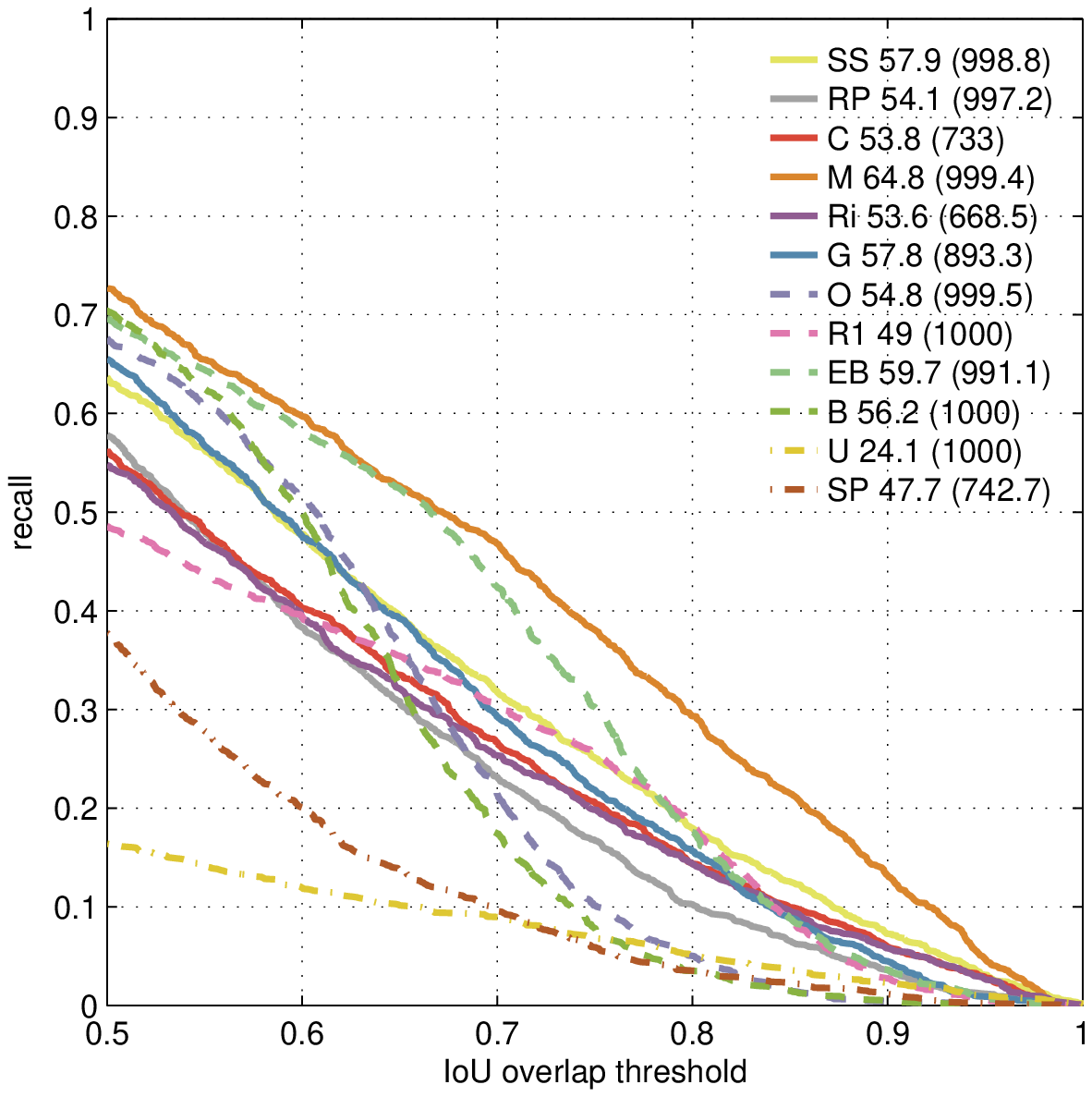}&
			\includegraphics[width=.2\linewidth]{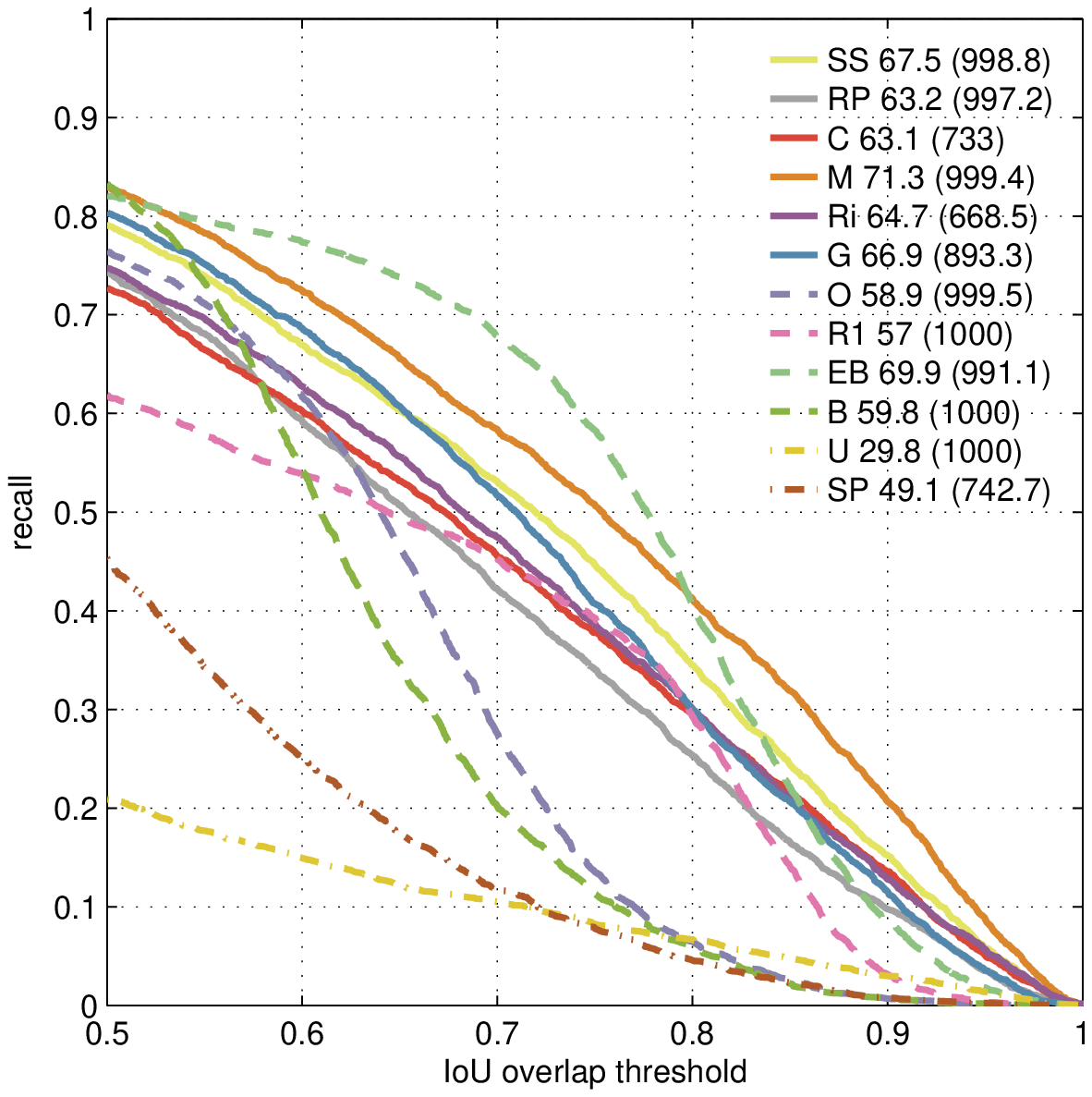}&	\includegraphics[width=.2\linewidth]{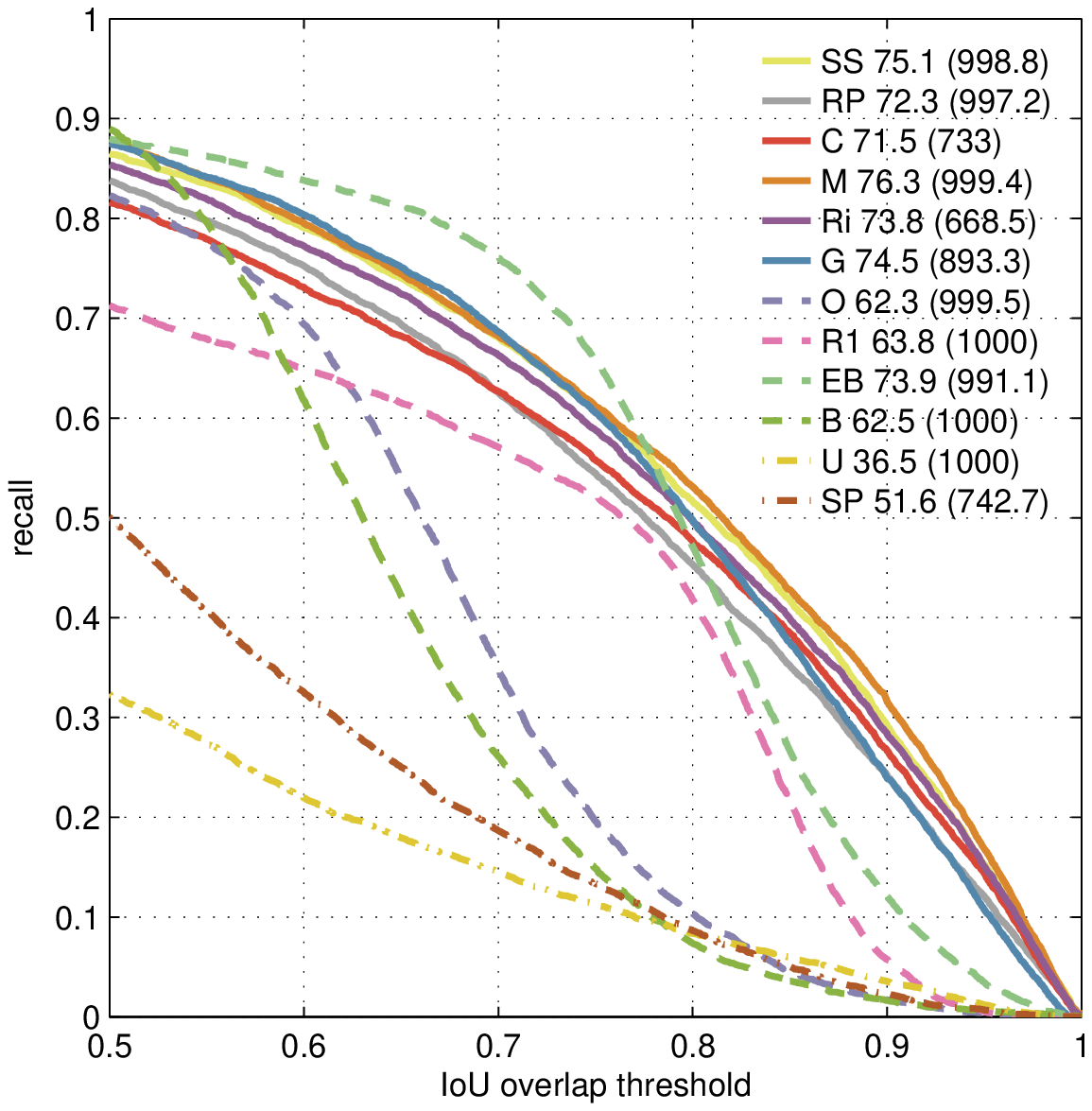}&
			\includegraphics[width=.2\linewidth]{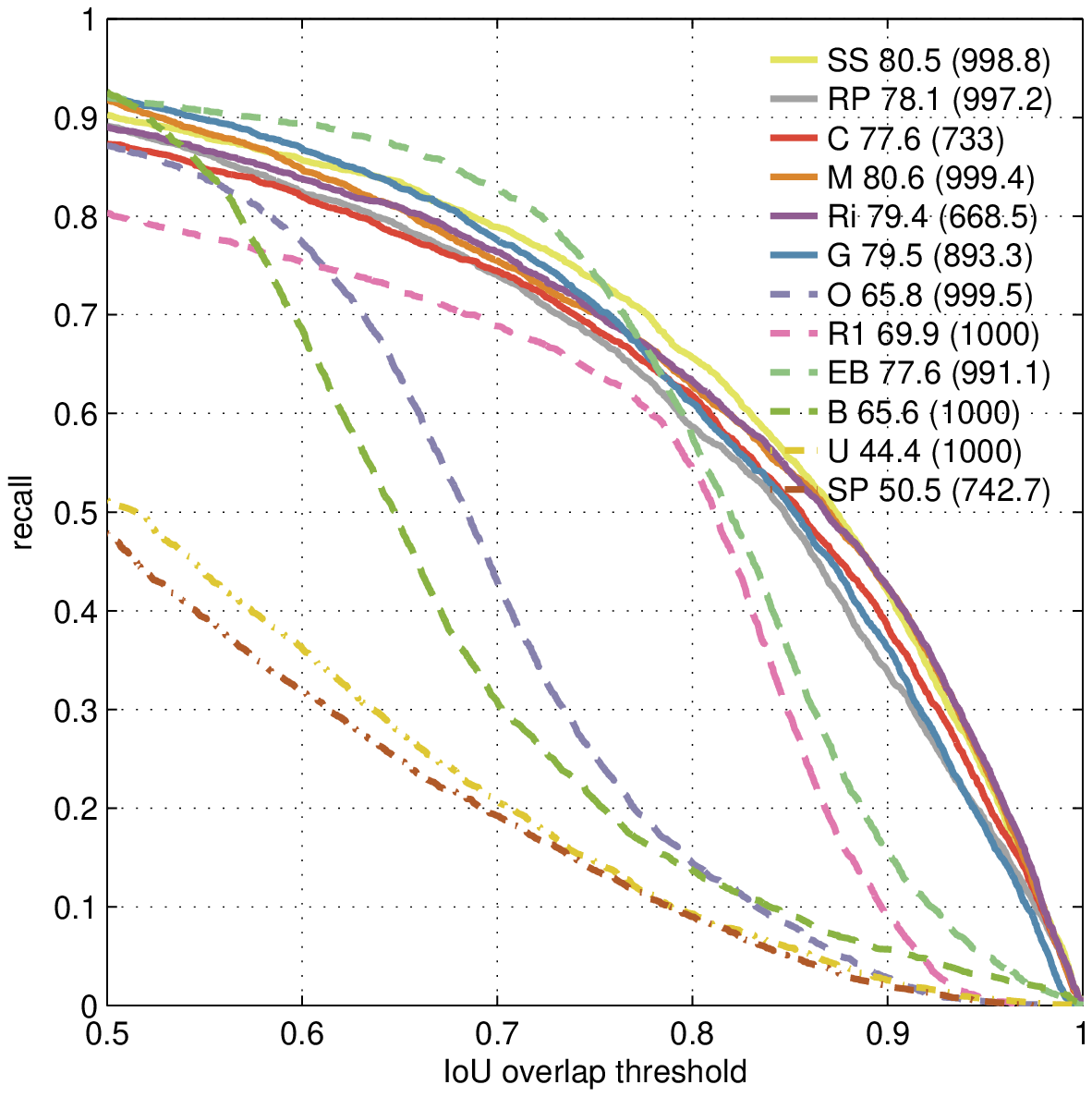}&
			\includegraphics[width=.2\linewidth]{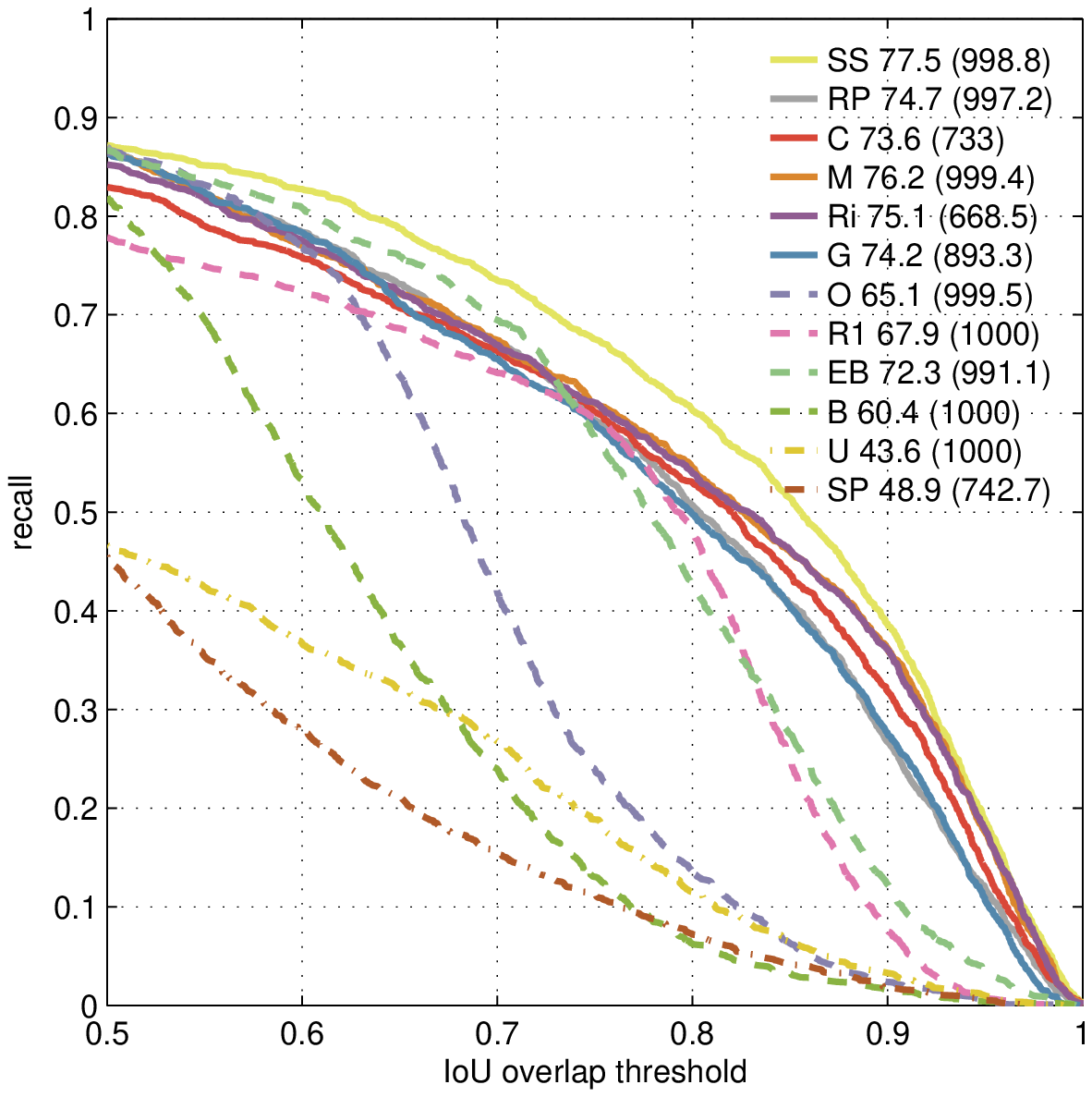}\\
			\tiny(a)&\tiny(b)&\tiny(c)&\tiny(d)&\tiny(e)
		\end{tabular}
		\vspace{3pt}
		\caption{\footnotesize Recall versus IoU threshold curves for different aspect ratio: \textit{`extra-tall'} (a), \textit{`tall'} (b), \textit{`medium'} (c), \textit{`wide'} (d) and \textit{`extra-wide'} (c). Dashed lines are for window based methods and solid lines are for region based methods. Dashed lines with dots are for the baseline methods.}
		\label{fig:iou_recall_1000_object_aspect}
	\end{minipage}
	\vspace{-20pt}
\end{figure*}

\textbf{Object Size} is measured as the pixel area of the bounding box, each bounding
box is assigned to a size category depending on the percentile sizes within the object instances: 
`extra-small' (XS: bottom 10\%); `small' (S: next 20\%);
`medium' (M: next 40\%); `large' (L: next 20\%); `extra-large' (XL: next 10\%). 
The ``recall versus IoU threshold'' curves for different object sizes
are shown in Figure ~\ref{fig:iou_recall_1000_object_size}. Existing methods exhibit various 
performance in terms of different sizes, typically
for objects from extra-small to medium size. EdgeBox shows the best performance for small to
medium range objects when $IoU \leq 0.7$. BING's performance is also stable across this size range. 
For other sizes, EdgeBox also shows comparable performance with the region based approach. Rahtu2011 (R1) is most size-sensitive window based method, which works decently for localizing large and extra-large objects. Such phenomenon for window based methods is correlated with the initial window sampling techniques (as shown in the the `uniform' baseline), which favour searching sizeable object first. 
MCG is the best performing region based method across different sizes. Interestingly, superpixel
baseline shows better performance than all existing methods for extra-small cases, which supports our previous hypothesis
that the pipeline of existing methods has a tendency to ignore small to medium scale objects or has difficulty in picking them out, though more component-wise analysis is left for future work.

\textbf{Aspect Ratio} is defined as the object width divided by height. The objects
are categorized into `extra-tall' (XT), `tall' (T), `medium' (M), `wide' (W) and `extra-wide' (XW), also
according to the percentile size within the dataset. The ``recall versus IoU threshold'' 
curves for different aspect ratios are shown in Figure ~\ref{fig:iou_recall_1000_object_aspect}.
Most objects from `medium' (M) to `wide' (W) are easier to localize than other aspect ratios, as many
 `extra-tall' and `tall' are correlated with `bottle' and `person' classes.
Existing methods experience variations in terms of various aspect ratios.
MCG and EdgeBox are the top performing methods when $IoU \leq 0.7$. The performance of region based methods
are correlated with the superpixel baseline, as superpixels tend to oversegment image
into regular shape regions. Another notable phenomenon is the performance variation 
for seeded region based approach (CPMC and RIGOR). 
One potnetial explanation is that CPMC and RIGOR applies
a regular grid based seeding approach, which may not produce stable seeding for elongated
objects. MCG and Geodesic applies an irregular seeding approach from contour map, which is more stable for objects in various aspect ratio. The performance of window based methods is also correlated
with initial window sampling, which can be observed by inspecting the `uniform'
baseline. The window based methods maintain performance at middle aspect ratio range, while show lower performance at two ends of the aspect ratio, 
due to the insufficient parameter sampling. 
\begin{figure*}[h]
	\begin{minipage}[b]{1\linewidth}	
		\centering
		\setlength{\tabcolsep}{0.5pt}
		\renewcommand{\arraystretch}{0.5}
		\begin{tabular}{ccc}
			\includegraphics[width=.3\linewidth]{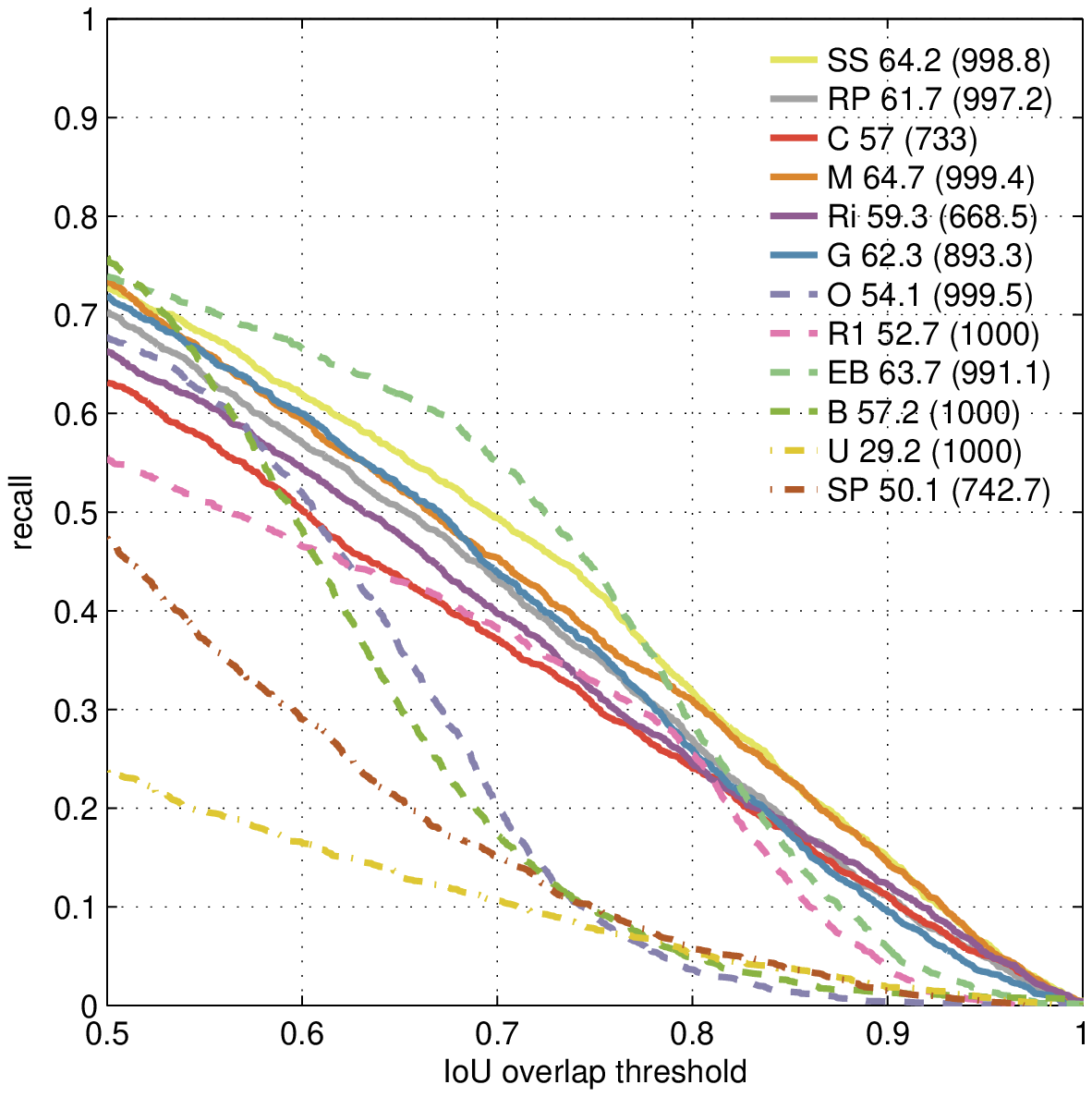}&
			\includegraphics[width=.3\linewidth]{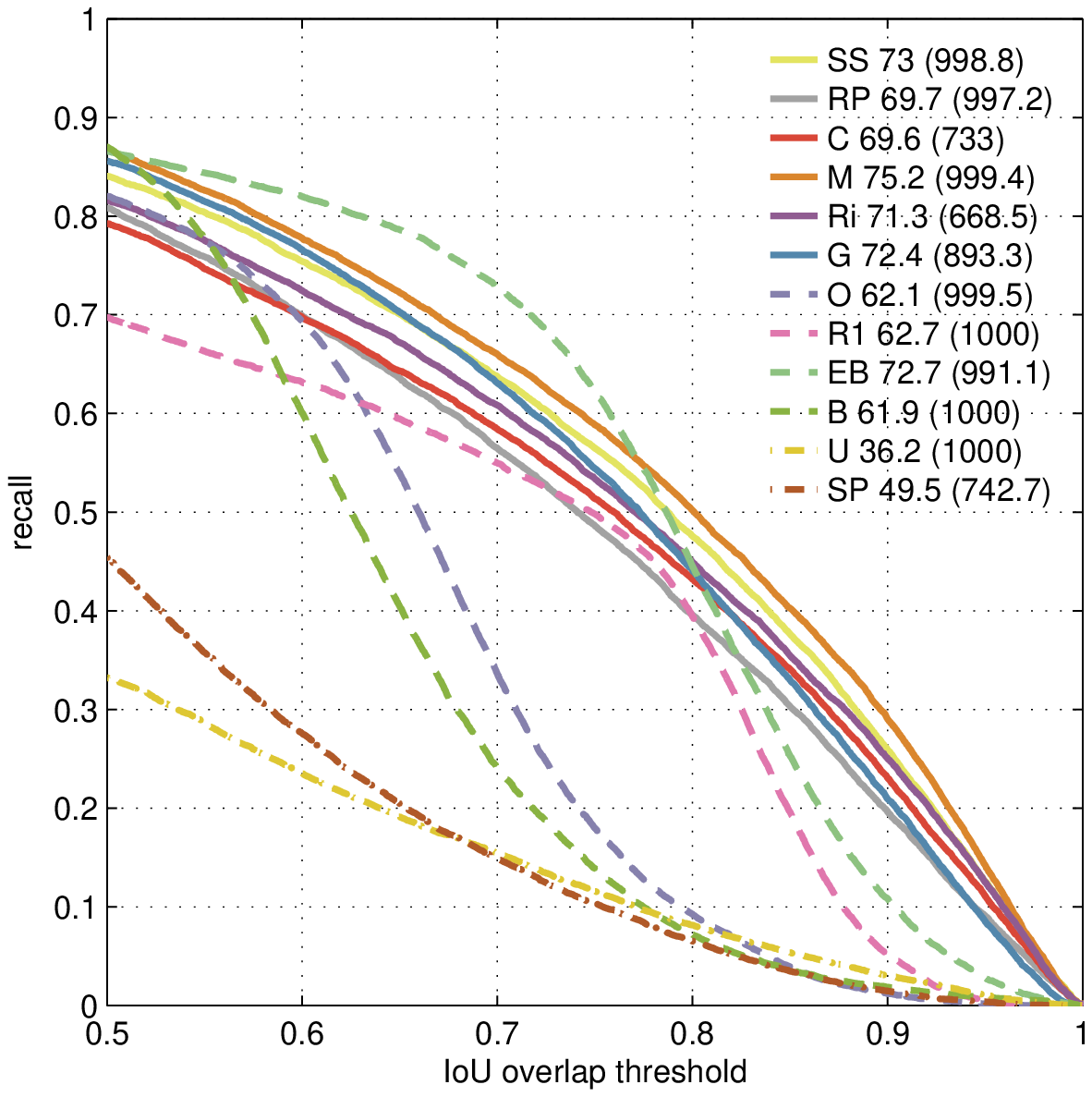}&
			\includegraphics[width=.3\linewidth]{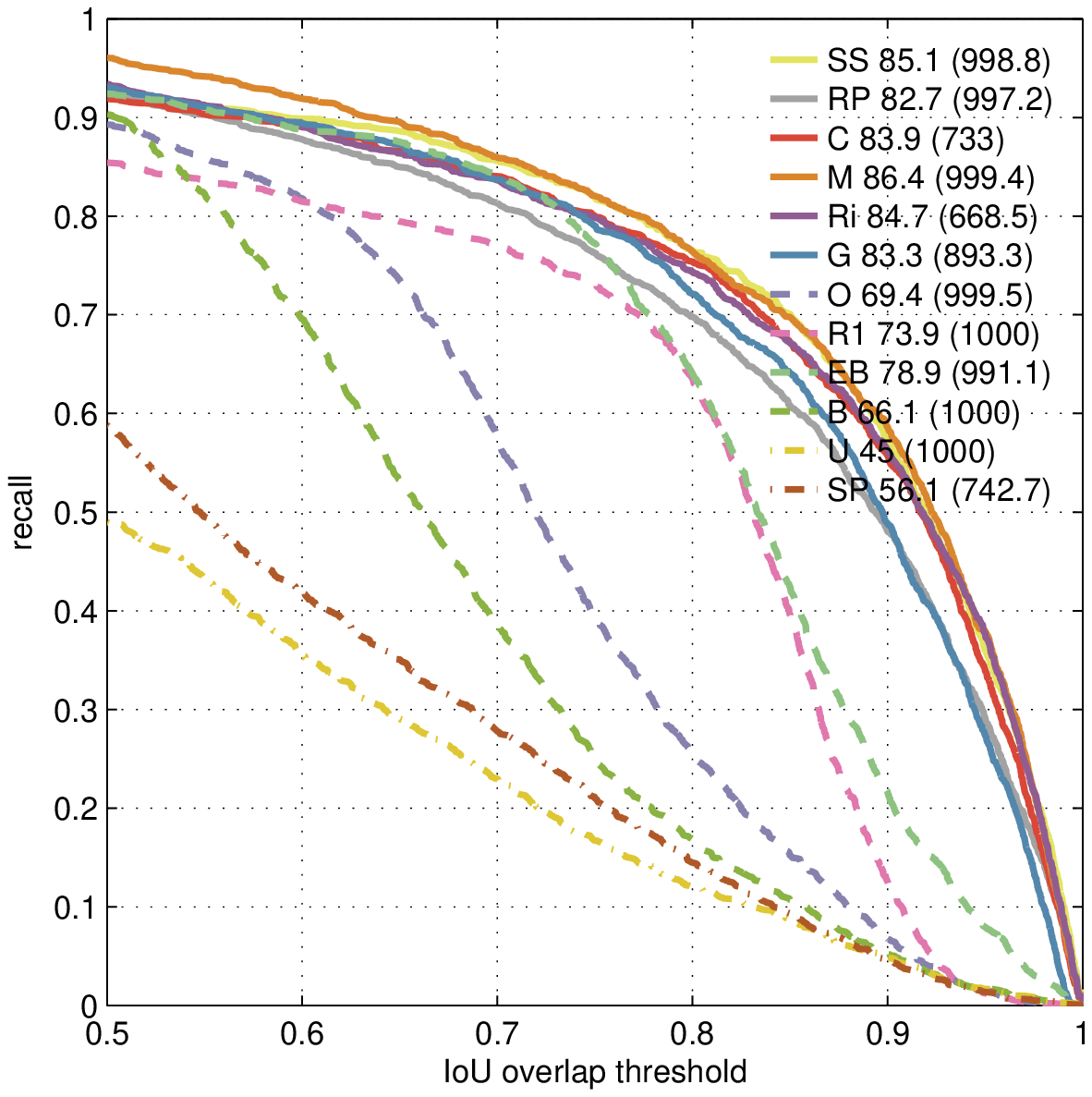}\\
			\tiny(a)&\tiny(b)&\tiny(c)
		\end{tabular}
		\vspace{3pt}
		\caption{\footnotesize Recall versus IoU threshold curves for different color contrast level: \textit{`low'} (a), \textit{`medium'} (b) and \textit{`high'} (c). Dashed lines are for window based methods and solid lines are for region based methods. Dashed lines with dots are for the baseline methods.}
		\label{fig:iou_recall_1000_color}
	\end{minipage}
	\\
	\begin{minipage}[b]{1\linewidth}	
		\centering
		\setlength{\tabcolsep}{0.5pt}
		\renewcommand{\arraystretch}{0.5}
		\begin{tabular}{ccc}
			\includegraphics[width=.3\linewidth]{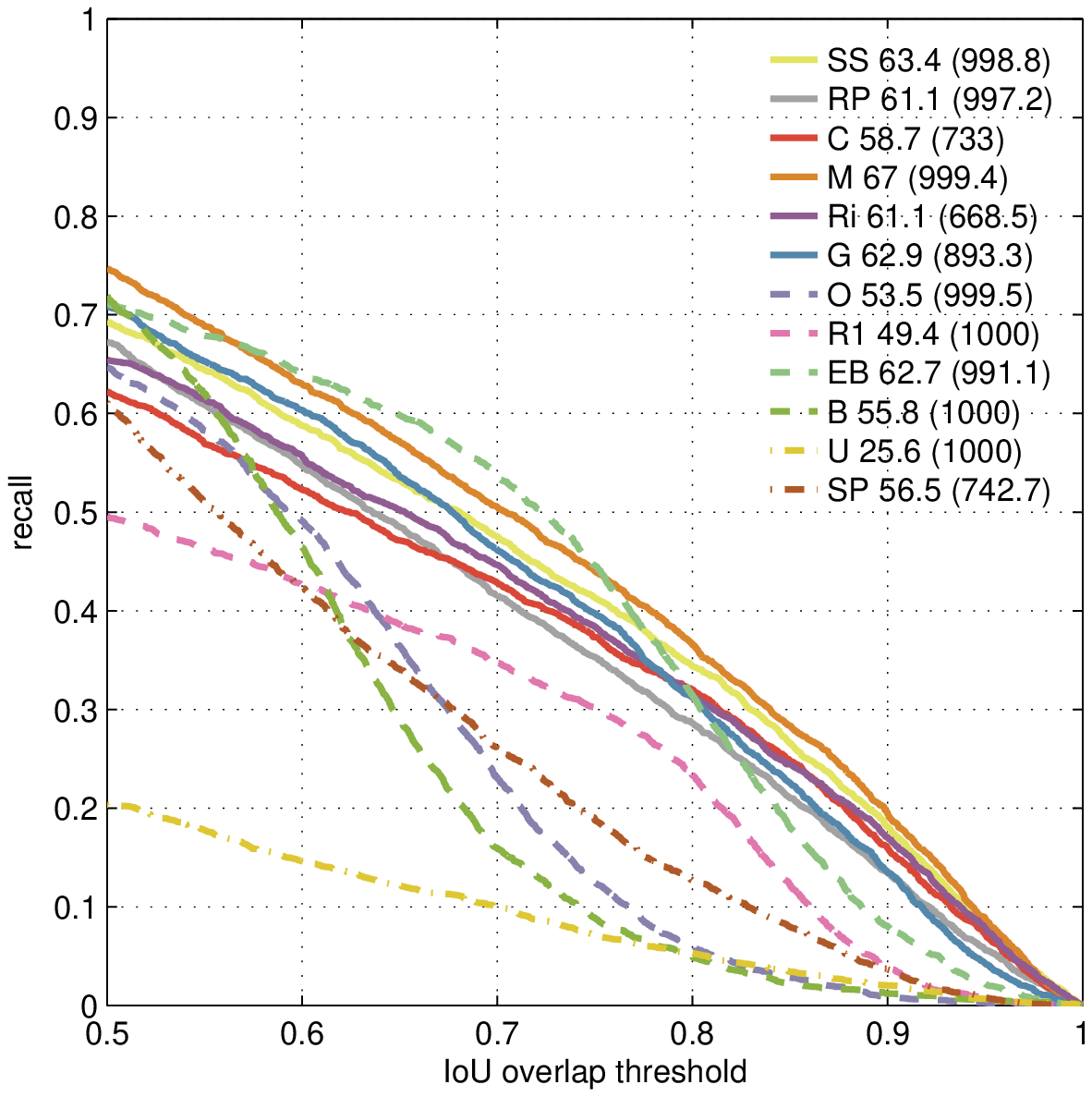}&
			\includegraphics[width=.3\linewidth]{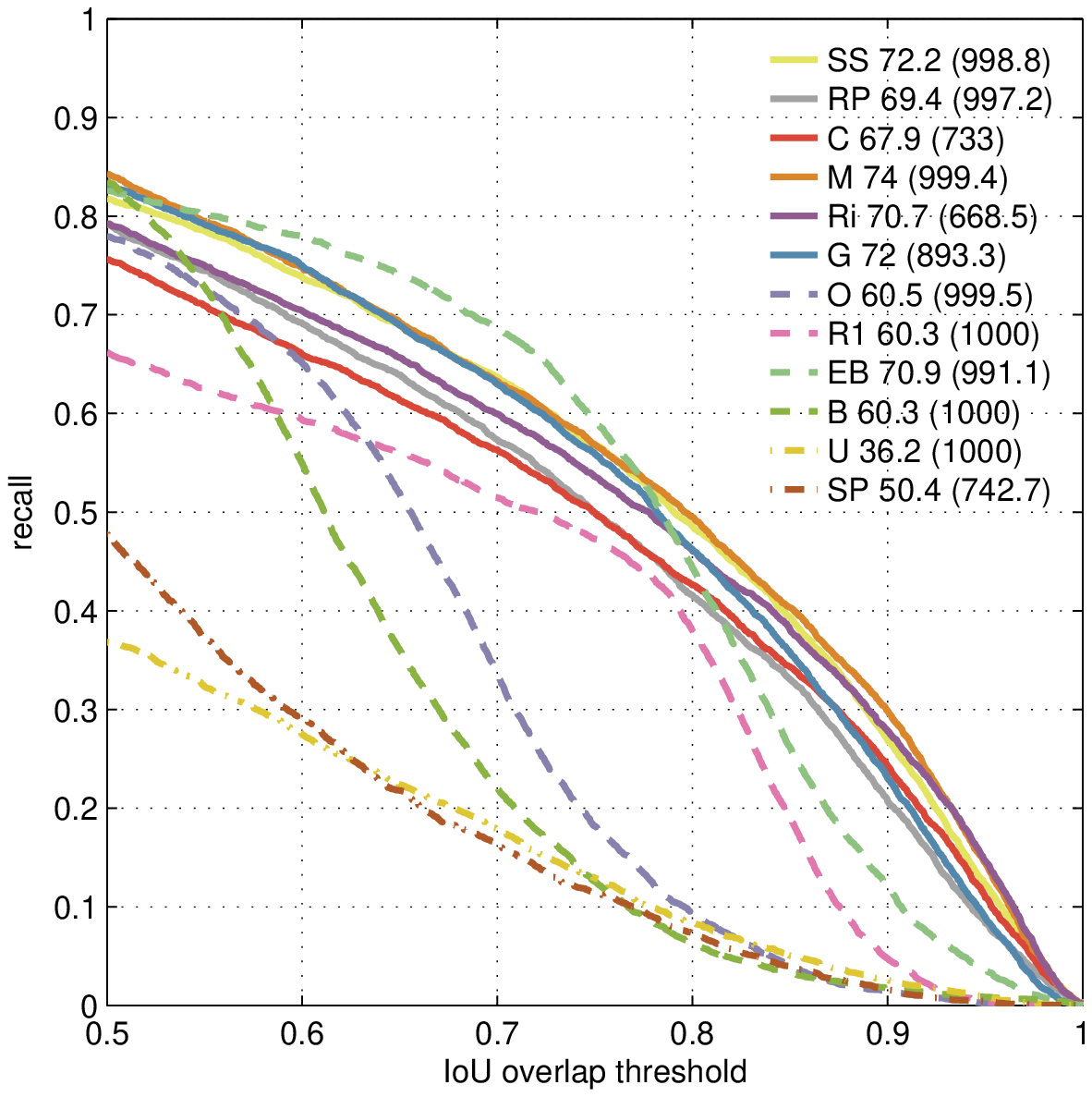}&		
			\includegraphics[width=.3\linewidth]{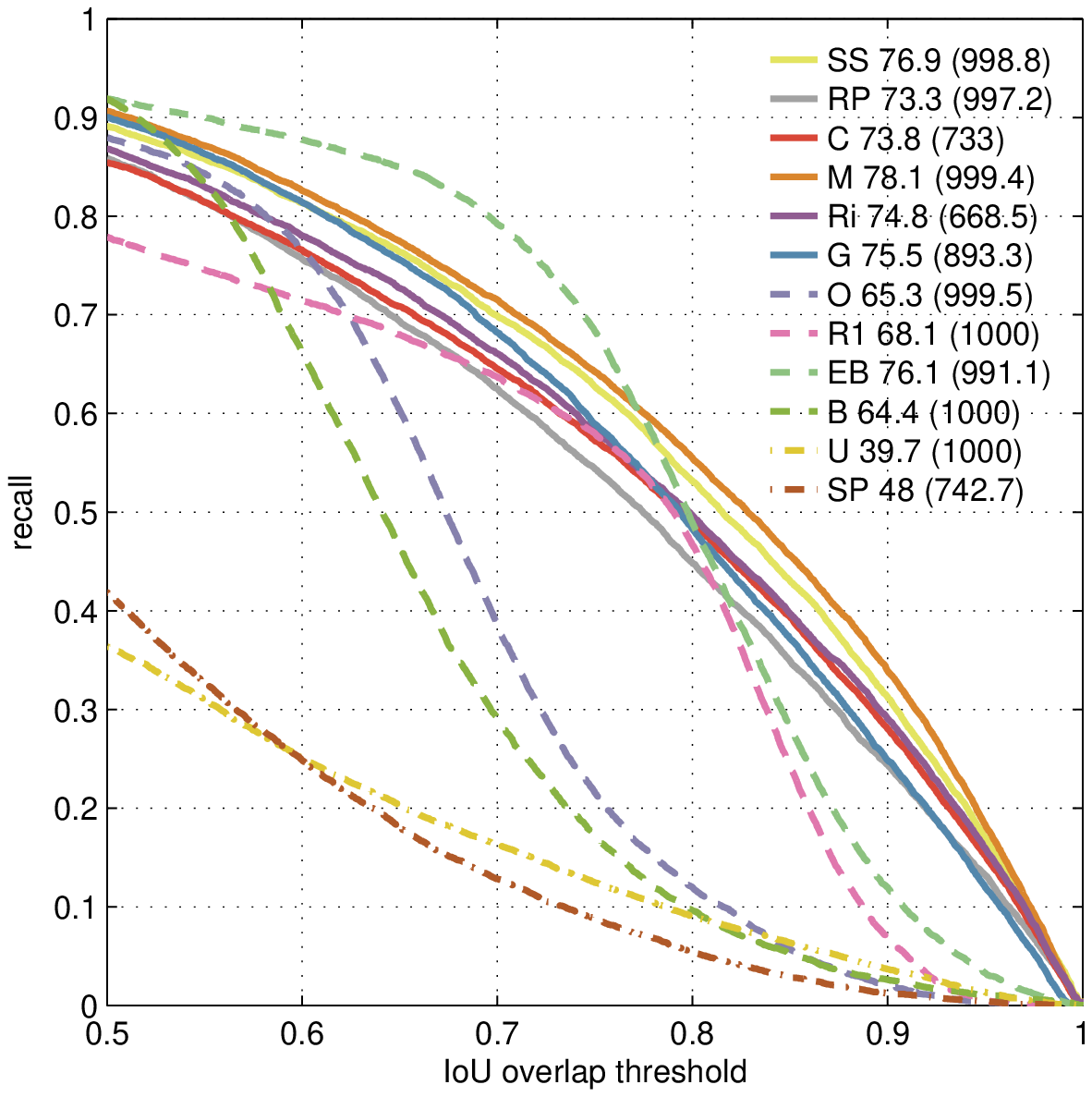}\\
			\tiny(a)&\tiny(b)&\tiny(c)
		\end{tabular}
		\vspace{3pt}
		\caption{\footnotesize Recall versus IoU threshold curves for different shape regularities: \textit{`low'} (a), \textit{`medium'} (b) and \textit{`high'} (c). Dashed lines are for window based methods and solid lines are for region based methods. Dashed lines with dots are for the baseline methods.}
		\label{fig:iou_recall_1000_shape}
	\end{minipage}
	\vspace{-25pt}
\end{figure*}

\textbf{Color Distinctiveness} is divided into three classes:  `low' (e.g. black car moving in the night), 
`medium' (e.g a pedestrian in human class), `high' (e.g. white horse on the grassland).
One common assumption for object proposal is that objects contain properties which help
them stand out from the background areas. Therefore, objects with strong contrast with the
background should be easier to get localized. The ``recall versus IoU threshold'' 
curves for different color contrast level are shown in Figure ~\ref{fig:iou_recall_1000_color}.
Our analysis did suggest such preference for highly contrast objects for existing methods, since many existing methods either explicitly modelling color contrast (e.g. CPMC, RIGOR, SelectiveSearch, Objectness) or implicitly making use of color contrast based features (e.g. MCG, EdgeBox, Geodeisc, RandomizedPrime etc.). A direct consequence for such design flavour is that region based methods are the most sensitive to color contrast changes, whereas window based methods suffer less because they benefit from gradient and shape cues (e.g. EdgeBox, BING and Objectness). 
On the other hand, Rathu2011 is quite sensitive to contrast changes as it relies on superpixel based window sampling technique.
As color distinctiveness is also
correlated with non-iconic objects, we evaluate the color contrast's influence in terms of
iconic and non-iconic views individually. The improvement shifts from low contrast to hight 
contrast for both kinds of objects follows similar performance improvement pattern (the figures can be found in last section of the supplementary material). Therefore, color contrast is an important factor in influencing existing state-of-the-arts. 

\textbf{Shape Regularity} is classified into three categories including `low' (e.g. tv monitor, bus), `medium' (e.g car), `high' (e.g. horse, dog). The ``recall versus IoU threshold'' curves for different shape regularities are shown in Figure ~\ref{fig:iou_recall_1000_shape}.
Nearly all existing methods make use of some shape or gradient information, such as BING, EdgeBox et al. Therefore,
it is interesting to inspect whether objects with more irregular shapes are easier to be located. 
Our experimental results partially support such intuition and are also consistent with
the method design. For example, EdgeBox directly evaluates the enclosed contour strengths.
The more irregular the shape enclosed, the more contour that may be included, and so
makes it easier to be localized. On the other hand, BING is less
sensitive to shape regularity change as it rely on a compactly represented shape
template. Region based methods are more sensitive to shape regularity changes, as they also relies on contour detections. Moreover, when
methods overly rely on such cues, they face difficulties under weak
illumination changes or other image degradations where the shape cues can be easily contaminated. This is supported by 
evaluating the shape regularity's influence in terms of
iconic and non-iconic view individually. The shape information carried non-iconic view
is weak, therefore the performance gap between these two kinds of objects is large (the figures can be observed
from last section of the supplementary material). 

\begin{figure}[h]
	\begin{minipage}[b]{1\linewidth}	
		\centering
		\setlength{\tabcolsep}{0.5pt}
		\renewcommand{\arraystretch}{0.5}
		\begin{tabular}{ccc}
			\includegraphics[width=.3\linewidth]{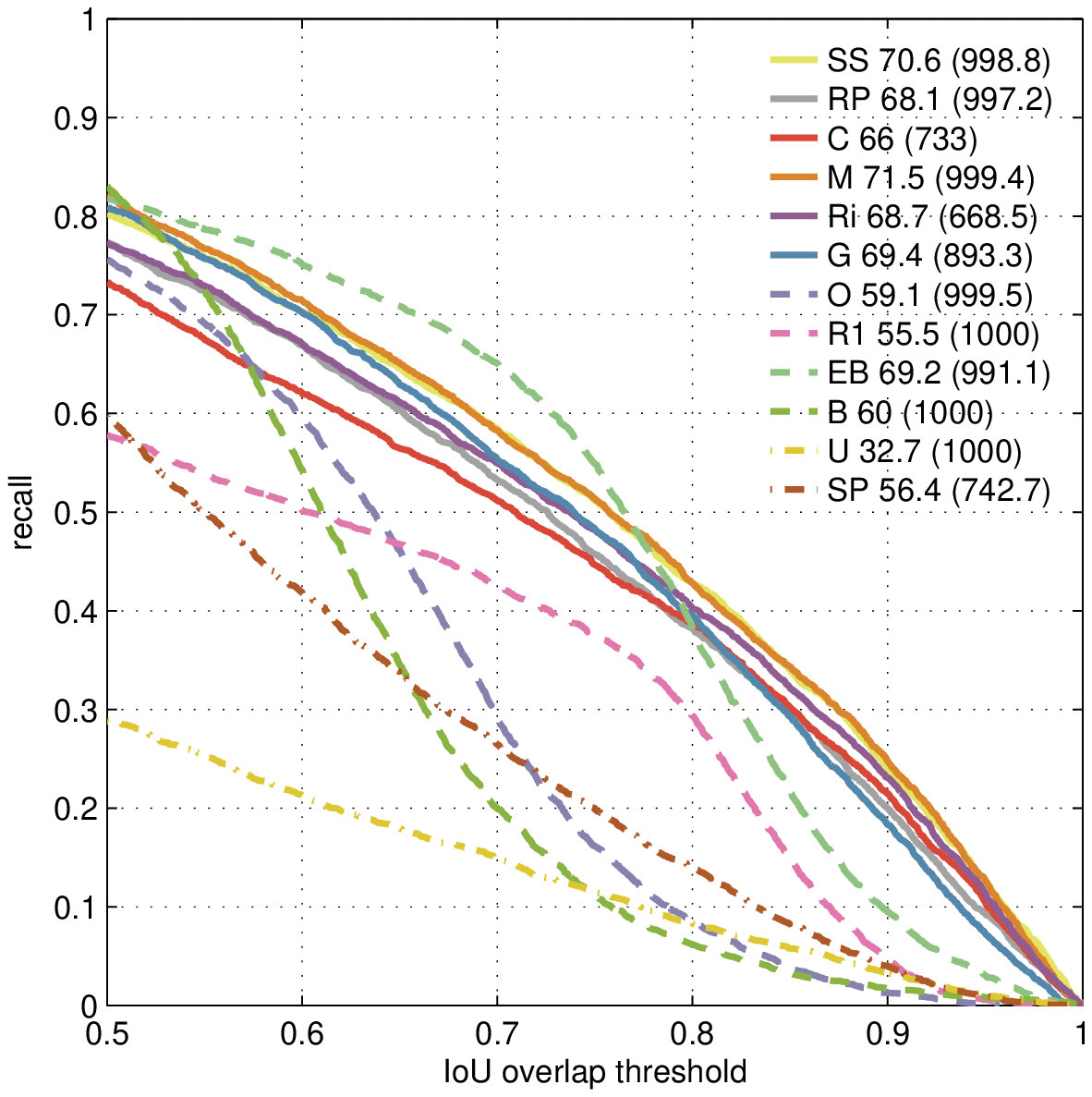}&
			\includegraphics[width=.3\linewidth]{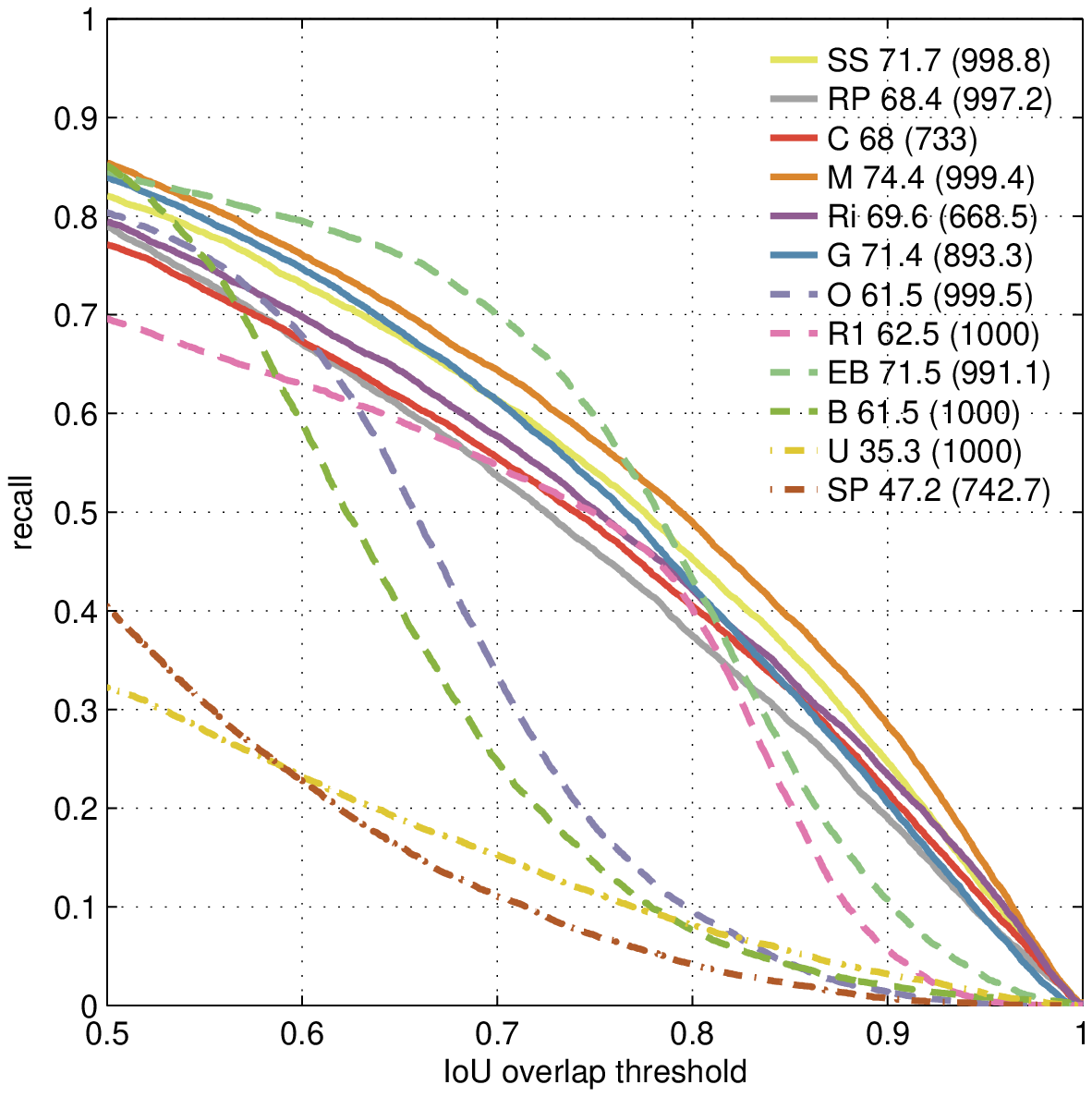}&		
			\includegraphics[width=.3\linewidth]{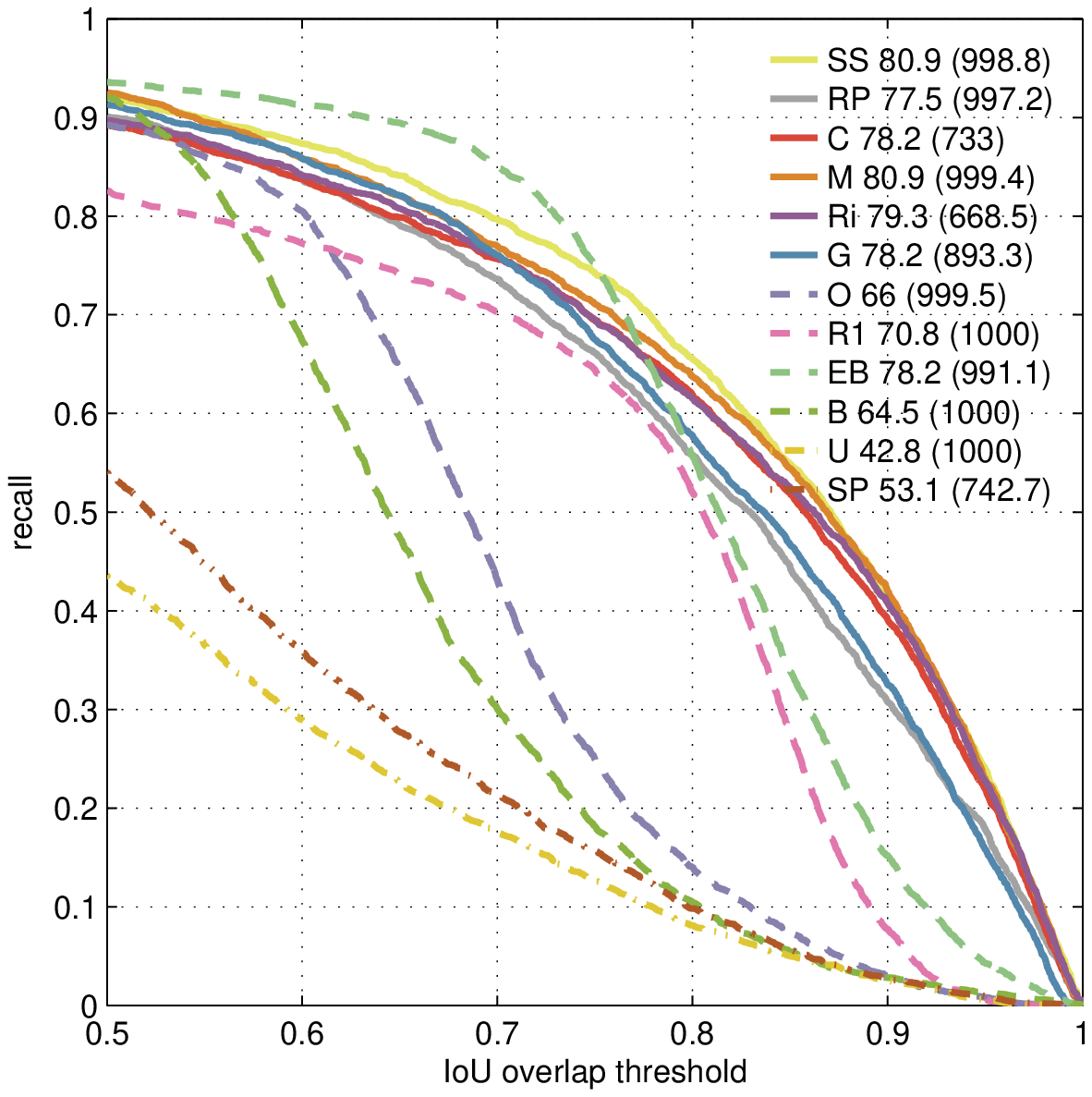}\\
			\tiny(a)&\tiny(b)&\tiny(c)
		\end{tabular}
		\vspace{3pt}
		\caption{\footnotesize Recall versus IoU threshold curves for different texture levels: \textit{`low'} (a), \textit{`medium'} (b) and \textit{`high'} (c). Dashed lines are for window based methods and solid lines are for region based methods. Dashed lines with dots are for the baseline methods.}
		\label{fig:iou_recall_1000_texture}
	\end{minipage}
	\vspace{-30pt}
\end{figure}

\textbf{Textureness} is defined by following classes: `low' (e.g. horse), `medium' (e.g sheep), `high' (e.g. damaitting).
The ``recall versus IoU threshold''  curves for different texture levels are shown in Figure ~\ref{fig:iou_recall_1000_texture}. According to the figures, there is little correlation between the performance and 
the amount of textures, except for Rathu. We conjecture that this phenomenon is caused by the spatial support and
feature design. Existing methods make use of large spatial support (e.g. bounding box and superpixels) to extract mid-level features and produce candidate regions, which is more robust to small scale texture change. 
Noted, single superpixel segmentation is not that repeatable across
different texture levels as indicated in the figure, existing region based approach makes use of diversified superpixel generation to greatly alleviates the sensitivity to textureness change.
\vspace{-20pt}
\section{Conclusion}
In this work, we study the influence of object-level characteristics over existing object 
proposal methods. We study existing methods' 
localization accuracy and latency, which suggests existing methods' performance variation for
different classes and their complementariness. Our further analysis along different
object characteristics suggest existing methods' pros and cons. BING, Geodesic,
EdgeBox, SelectiveSearch and MCG are the methods which achieve stable performance across
different characteristics. In terms of localization accuracy,
MCG is the best region based method across various properties, though the speed is mediocre. 
The speed of BING is good, however its localization is quite poor. On the other hand,
EdgeBox, Geodesic, SelectiveSearch achieve good balance between accuracy and speed. 
A more detailed summarization along each dimensions can be found in supplementary material.

The study of iconic view reveals existing methods' 
difficulty in localizing objects outside canonical perspectives, hence
future work should focus more on improving the recall rate at non-canonical view
and reducing the influence of location prior. This also applies to experiment and 
dataset design. Our concerns is that further 
object proposal research along PASCAL VOC and ImageNet has a risk of over-fitting
the large numbers of iconic objects but achieves little improvements for non-iconic objects. The recent MSCOCO
dataset~\cite{COCO} is a good attempt by collecting images with more
balanced distribution in perspective change. 

Existing methods also have limitations in terms of object size, aspect ratio, color contrast and shape change. The improvement with respect to these attributes could be substantial though it may not be an easy task as difficult cases are combinations of different challenges. Large improvement in a few aspects may bring
small improvement in terms of overall performance, therefore further analysis along different attributes is encouraged to better reflect the performance development. 

There are some potential
directions worthwhile for exploration, e.g. developing specific template 
for challenge cases such as low contrast, illumination and resolution~\cite{ParkRF10}. Further gain
can be can also be achieved by using context~\cite{Torralba03}. Moreover, most existing methods use
bottom-up methods to generate object proposal, and
the combination with mid-level shape cues can be helpful 
as investigated in ShapeSharing~\cite{ShapeSharing}. 

In our future work, we will conduct
similar research for MSCOCO, which has a different distributions from PASCAL VOC
and includes more annotated attributes.  Such work will further reveal
insight of existing methods and cast light for future research. 
 
\section{Acknowledgement}
The project is funded by Office for Space Technology and Industry (OSTIn) Space Research Program under Economic Development Board, Singapore. The project reference no is S14-1136-IAF OSTIn-SIAG.

\bibliography{egbib}

\end{document}